%% file: main.tex
\documentclass[10pt]{article}
\usepackage[preprint]{tmlr}

\usepackage{fancyhdr}
\renewcommand{\headrulewidth}{0pt} 
\fancypagestyle{firstpage}{
    \renewcommand{\headrulewidth}{0pt}
}

\usepackage{graphicx} 
\usepackage{cuted}
\usepackage{float}
\usepackage{framed,color}
\usepackage{caption}
\usepackage{bm} 
\usepackage{amsmath}
\usepackage{amssymb}
\usepackage{booktabs}
\usepackage{multirow}
\usepackage{graphicx}
\usepackage{subcaption}
\usepackage{cite}
\usepackage{xcolor}

\usepackage[hidelinks]{hyperref}
\definecolor{citecolorcustom}{RGB}{124,109,155}
\hypersetup{
  colorlinks=true,
  citecolor=citecolorcustom,  
  linkcolor=citecolorcustom,
  urlcolor=citecolorcustom
}

\newcommand{\eg}{\emph{e.g.}}
\newcommand{\ie}{\emph{i.e.}}
\newcommand{\intern}{\dagger}
\newcommand{\equal}{*}
\newcommand{\corres}{\ddag}

\title{
\centering
Scalable Vision-Language-Action Model Pretraining for \\
\!Robotic \kern-0.1em Manipulation \kern-0.1em with \kern-0.1em Real-Life \kern-0.1em Human Activity \kern-0.1emVideos}

\author{
\centering
    Qixiu Li$^{1,2\equal\intern}$ \ 
    Yu Deng$^{2\equal}$  \ 
    Yaobo Liang$^{2\equal}$ \ 
    Lin Luo$^{2\equal}$ \ 
    Lei Zhou$^{2\intern}$ \ 
    Chengtang Yao$^{2}$ \ 
    \\
    Lingqi Zeng$^{2\intern}$ \ 
    Zhiyuan Feng$^{1,2\intern}$ \ 
    Huizhi Liang$^{1,2\intern}$ \ 
    Sicheng Xu$^{2}$ \ 
    Yizhong Zhang$^{2}$ \ 
    Xi Chen$^{2}$ \ 
    \\
    Hao Chen$^{2}$ \ 
    Lily Sun$^{2}$ \ 
    Dong Chen$^{2}$ \ 
    Jiaolong Yang$^{2\ddag}$ \ 
    Baining Guo$^{2}$
    \\
    \normalfont
    \vspace{4pt}
	$^1${Tsinghua University} \quad $^2${Microsoft Research Asia}\\ 
    \vspace{4pt}
    \url{https://microsoft.github.io/VITRA/}
}

\begin{document}
\thispagestyle{firstpage}

\vspace*{15pt}
\maketitle

\begingroup
\renewcommand\thefootnote{}
\footnotetext{
$^{\equal}$ Equal contribution. $^{\intern}$ Intern work done at Microsoft Research Asia. $^{\corres}$ Corresponding author.}
\addtocounter{footnote}{0}
\endgroup

\input{sections/0_abstract}
\input{sections/1_intro}
\input{sections/2_related}
\input{sections/3_method}

\input{sections/4_experiment}

\input{sections/5_discussion}

\section{Conclusion}
This paper introduces a novel approach for pretraining robotic manipulation VLA models using unstructured real-life human activity videos. We develop a fully-automatic pipeline to convert in-the-wild egocentric human videos into atomic-level VLA data aligned with existing robotic demonstrations. We also design  a dexterous hand VLA model with tailored training strategies to effectively leverage human data for pretraining. Experiments show that our pretrained model exhibits strong zero-shot performance in unseen real-world environments, high task success after being finetuned on limited robot data, and favorable data scaling behavior, demonstrating a highly promising and scalable approach toward learning truly generalizable embodied robots.

\section*{Acknowledgments}
We would like to thank Mozheng Liao, Guanghao Wang, and Bo Liang for their help in building the hardware system, Lidong Zhou for the suggestions on analysis experiments, and Yunze Liu, Ruicheng Wang, Fangyun Wei, Yichao Shen, and Jianmin Bao for discussions on improving this work.

\bibliography{main}
\bibliographystyle{plainnat}

\clearpage
\appendix

\renewcommand{\thefigure}{\Roman{figure}}
\renewcommand{\thetable}{\Roman{table}}
\renewcommand{\theequation}{\Roman{equation}}
\setcounter{section}{0}
\setcounter{figure}{0}
\setcounter{equation}{0}
\input{sections/suppl}

\end{document}

%% file: sections/0_abstract.tex
\begin{abstract}
This paper presents a novel approach for pretraining robotic manipulation Vision-Language-Action (VLA) models using a large corpus of unscripted real-life video recordings of human hand activities. 
Treating human hand as dexterous robot end-effector, we show that ``in-the-wild'' egocentric human videos without any annotations can be transformed into data formats fully aligned with existing robotic  V-L-A training data in terms of task granularity and labels.  
This is achieved by the development of a fully-automated holistic human activity analysis approach for arbitrary human hand videos. This approach can generate atomic-level hand activity segments and their language descriptions, each accompanied with framewise 3D hand motion and camera motion. We process a large volume of egocentric videos and create a hand-VLA training dataset containing 1M episodes and 26M frames. This training data covers a wide range of objects and concepts, dexterous manipulation tasks, and environment variations in real life, vastly exceeding the coverage of existing robot data.
We design a dexterous hand VLA model architecture and pretrain the model on this dataset. The model exhibits strong zero-shot capabilities on completely unseen real-world observations. Additionally, fine-tuning it on a small amount of real robot action data significantly improves task success rates and generalization to novel objects in real robotic experiments. We also demonstrate  the appealing scaling behavior of the model's task performance with respect to pretraining data scale.
We believe this work lays a solid foundation for scalable VLA pretraining, advancing robots toward truly generalizable embodied intelligence.

\end{abstract}

%% file: sections/1_intro.tex
\section{Introduction}

Pretraining on large, generic data is the key for models to acquire commonsense knowledge and achieve domain generalization. While the pretraining of Large Language Models (LLM) and Vision Language Models (VLM) has seen remarkable success~\citep{brown2020language,liu2023visual,achiam2023gpt,team2023gemini}, pretraining Vision-Language-Action (VLA) models for dexterous hand manipulation remains largely underexplored. 

\begin{figure}[!t]
\vspace{3pt}
	\centering
\includegraphics[width=1.0\textwidth]{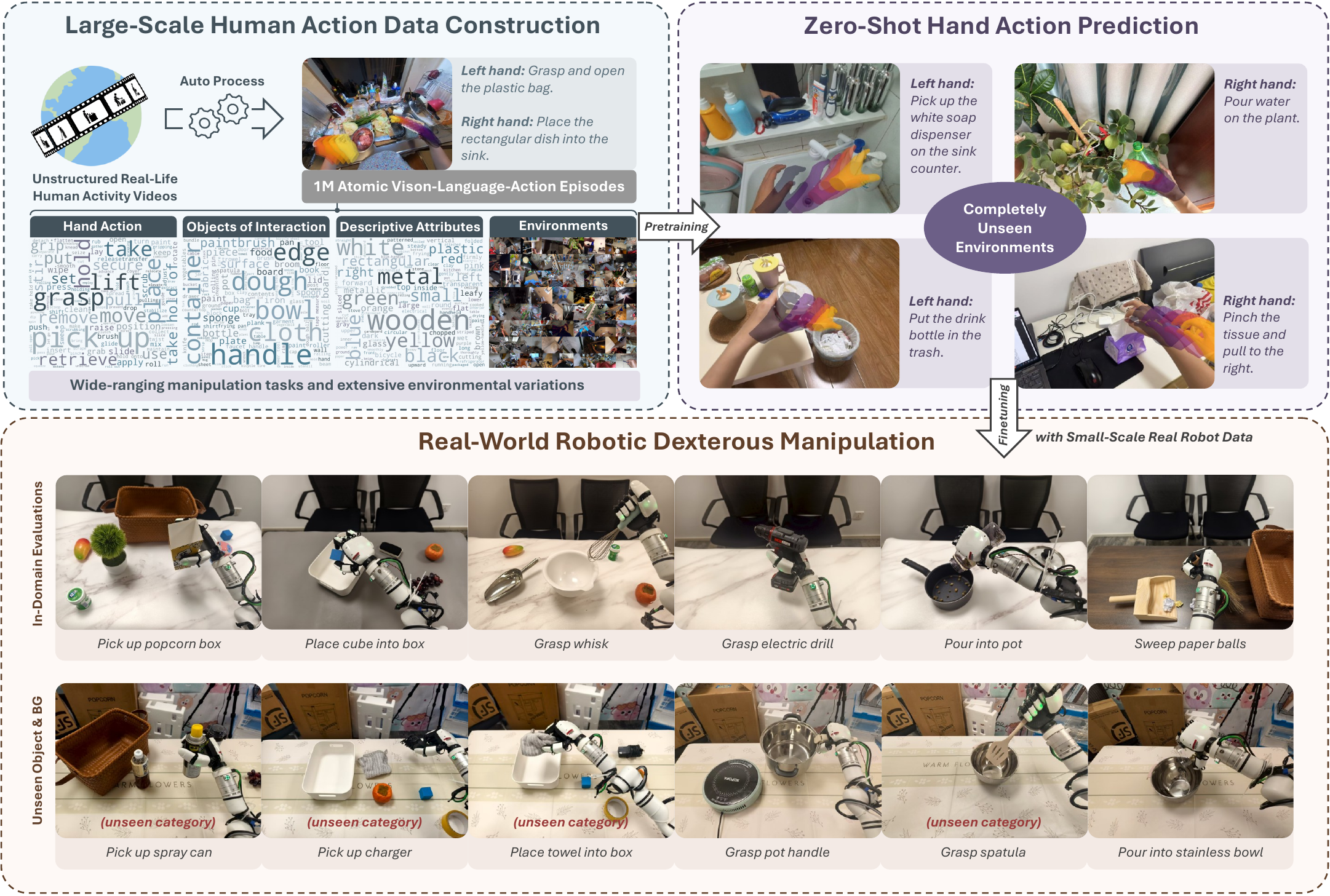}
\vspace{-17pt}
	\captionof{figure}{We present a pretraining approach for robotic Vision-Language-Action (VLA) models by transforming unstructured real-life videos of human activity into structured V-L-A formats aligned with existing robot data. The pretrained model demonstrates strong zero-shot hand action prediction in unseen environments and can be effectively fine-tuned with dexterous robot hand data for real-world tasks, showing robust generalization to new objects and environments.}	
	\label{fig:teaser}
\vspace{-5pt}
\end{figure}

Existing Vision-Language-Action data for robotic manipulation are typically collected in laboratory settings through human teleoperations~\citep{zitkovich2023rt,o2024open,bu2025agibot,khazatsky2024droid,fang2024rh20t}. Although such robot action data is invaluable, its high acquisition cost significantly limits both the scale of the collected data and its diversity in  skills, object categories, and scene variations. Consequently, current V-L-A datasets lag far behind the Internet-scale language and VL data in terms of quantity and diversity, and they fall short of representing the complexity required for real-world robotic tasks. The V-L-A data for dexterous robot hands is even more scarce; to our knowledge there are no large-scale dexterous hand action datasets available for pretraining.

Meanwhile, there is a vast amount of real-life human videos on the web, containing rich examples of everyday human actions and physical interactions with diverse environments.
These videos are typically \textbf{\emph{unstructured}}: they come unscripted and unsegmented, vary in length and task granularity, contain noisy and irrelevant actions, and lack language instruction and 3D action labels.
Although there have been numerous interests in utilizing human video for robot learning~\citep{qin2022dexmv,xiao2022masked,ma2023vip,nair2023r3m,wang2023mimicplay,shaw2023videodex,hu2025video,bahl2023affordances,yang2025magma,yelatent2025,chen2024igor,bjorck2025gr00t}, no existing approaches leverage  large-scale, unstructured videos without any human annotation for VLA model pretraining.
This leads to a critical question: \textbf{\emph{can we transform these unstructured videos into data formats fully aligned with existing robotic V-L-A training data?}}

This work is the first to address this question, and we provide an affirmative answer. For unstructured human videos, we consider human hand as robot end-effector and seek to achieve two types of alignment with real robot V-L-A data.
\emph{1) Task alignment}: we need meaningful segmentation and filtering of atomic-level human action sequences (short-horizon tasks), adhering to the recipe of existing robot data. This problem is closely related to temporal action segmentation from videos, which remains an open problem and there are no existing methods that meet our needs.
\emph{2) Label alignment}: we need to recover metric-space 3D hand motion accurately to the extent possible\footnote{While it's ideal to have 3D motion labels as accurate as possible, we believe some noise and imperfection are acceptable for pretraining, where the goal is to grasp common knowledge, learn motion patterns for diverse skills, and experience a wide spectrum of object and scene variations.} to provide dense action labels. This is difficult as we often work with single, uncalibrated, and likely moving cameras. 
Additionally, we need precise language instruction labels to describe the actions.

To this end, we introduce a holistic human activity analytic framework that converts any human hand activity video of arbitrary length into multiple V-L-A trajectories of dexterous manipulation. It is a fully-automatic approach requiring no human intervention. 
In this framework, we first develop a monocular 3D camera and hand pose tracking approach leveraging recent advancements in 3D vision community, particularly deep visual SLAM, depth estimation, and hand reconstruction. The outputs include the camera FoV, the framewise camera pose, and the framewise hand pose (based on the 6D wrist pose and full joint angles). For temporal atomic action segmentation, we propose a simple yet surprisingly effective algorithm based on the hand movement speed in the 3D space, obtained from the recovered 3D motion labels. Finally, for each segmented video clip, we visualize hand trajectories on sampled video frames and prompt VLM to determine whether the action constitutes meaningful manipulation and, if so, describe it in natural language.

\emph{One significant advantage of real-life video data is the inherent action diversity and scene variation it offers}. As a starting point, we process a large volume of raw videos from existing egocentric human video datasets.
The resultant hand V-L-A dataset contains about 1 million episodes and 26 million frames, This dataset captures a broad spectrum of objects, concepts, skills, and environmental variations, vastly exceeding the coverage of existing robot data as shown in our analysis. We also develop a dexterous hand VLA model architecture with a Causal Action Transformer and pretrain the model on this dataset. The model exhibits strong zero-shot capabilities on observations of completely new scenes, a level of performance not seen in any prior method. We conduct real-world robot experiments and show that fine-tuning the model on a small amount of real robot hand data significantly improves task success rates and generalization to novel objects and backgrounds. Furthermore, our experiments show \emph{a clear scaling behavior of task performance with respect to pretraining data scale.}

Our work stands distinct from prior research that utilizes human video for training robotic manipulation models. The approaches that leverage egocentric human video for learning vision and language representations, affordances, point trajectories, \emph{etc.}~\citep{xiao2022masked,ma2023vip,nair2023r3m,hu2025video,bahl2023affordances,yang2025magma}, did not explore action pretraining for VLA models. Recent works that use latent actions from human videos \citep{yelatent2025,bjorck2025gr00t,chen2024igor} for pretraining do not provide explicit 3D action labels as we do. Our experiments demonstrate the superiority of our pretraining approach. Most recently, a few works that are concurrent to ours studied training VLA models with explicit 3D hand motions similar to ours~\citep{yang2025egovla,luo2025being,bi2025h}, but their data is largely limited to scripted laboratory captures; a detailed discussion is provided in the next section.

\emph{Our approach offers a more tractable way for pretraining data scaling compared to existing techniques.}
Although this work uses videos from existing egocentric video datasets, there are no technical barriers preventing further data scaling.
By not imposing constraints on the subjects' activities or environments and requiring only a single webcam, every life recorder can effectively become a robot teacher. We envision a future where robots can effectively learn from abundant, low-cost human video demonstrations to acquire diverse skills, complemented by targeted fine-tuning using a modest amount of real robot data or reinforcement learning. 
{\emph{Our training dataset and pretrained VLA models will be open-sourced to the community to facilitate further research.}} 

%% file: sections/2_related.tex
\section{Related Works}

\newcommand{\paravspaceRelated}{\vspace{5pt}}

\paragraph{Robotic VLA Model Pretraining}
Robotic VLA models~\citep{brohan2022rt,team2024octo,li2022vision,zitkovich2023rt,o2024open,black2410pi0,li2024cogact,liu2024rdt,wen2025tinyvla,intelligence2025pi_,qu2025spatialvla,wen2025dexvla} that can perform diverse language-instructed tasks typically need pretraining on large data. 
Incorporating VL-pretrained modules or backbones has been a common practice for VLA models, and here we focus on a brief overview of the pretraining with regard to \emph{the action modality} or its proxy. Most recent VLA models with action pretraining~\citep{team2024octo,o2024open,li2024cogact,qu2025spatialvla,black2410pi0,intelligence2025pi_,kim2025fine} have leveraged the Open X-Embodiment (OXE) dataset~\citep{o2024open}, which contains over 1M real robot trajectories collected on over twenty robots. This large-scale dataset provides diverse skills and environment variations well suited for pretraining. Some of these works~\citep{liu2024rdt,qu2025spatialvla,black2410pi0,intelligence2025pi_} also incorporate more open-source or in-house robot action data in addition to OXE. The work of \citep{deng2025graspvla} synthesized a large volume of V-L-A data in simulators for pretraining, but it handles the grasping task only. A line of works~\citep{yelatent2025,chen2024igor,bu2025agibot,bjorck2025gr00t,bu2025learning} studied learning latent action from human and/or robot videos in an unsupervised manner and pretraining models using the extracted latents as the proxy for action. Some other works propose to use the future frames in videos as the prediction target for pretraining~\citep{wu2023unleashing,cheang2024gr}.

There are some recent attempts concurrent to ours which use  3D hand action labels of egocentric human videos for VLA pretraining~\citep{yang2025egovla,luo2025being,bi2025h}. They primarily  use hand-object interaction videos captured in controlled environments with privileged information. For example, the videos are well segmented to language-instructed action clips since the tasks are pre-scripted, and 3D hand motions are typically obtained with advanced devices (\emph{e.g.}, RGBD sensors, VR/AR headsets). We focus on a different goal, \emph{i.e.}, harnessing unscripted real-life human videos for large-scale pretraining, which encompass a significantly  broader range of tasks, objects, and real-world environments. This greatly enhances the zero-shot action perdition performance.
Furthermore, their casual capture nature facilitates much greater scalability.

\vspace{-4pt}
\paragraph{Dexterous Hand Manipulation} 
Dexterous manipulation with multi-fingered robot hands has been a vibrant area of research for decades. Earlier learning-based models with visual inputs were typically trained with reinforcement learning in simulators~\citep{akkaya2019solving,andrychowicz2020learning,chen2022towards}.  However, training dexterous RL policies  requires sophisticated reward design and their applicability in real-world scenarios is often limited. Using human teleoperated demonstrations for imitation learning was also widely used to improve task performance~\citep{rajeswaran2018learning,jain2019learning}. Methods that utilize human hand motion as demonstration data~\citep{handa2020dexpilot,qin2022dexmv,wang2023mimicplay,wang2024dexcap,qiu2025humanoid,kareer2025egomimic,park2025learning} were also actively studied. These previous works typically address a single or small set of tasks for a trained model. Recently, language instruction has been incorporated into dexterous hand manipulation models to handle more tasks with diverse objects~\citep{zhong2025dexgraspvla,hu2025video,de2025scaffolding}.

\vspace{-4pt}
\paragraph{Robot Learning from Human Videos}
Exploiting human videos to train robotic models has been actively studied in recent years.
Several studies~\citep{xiao2022masked,ma2023vip,nair2023r3m,yang2024spatiotemporal} leverage egocentric human videos for learning vision and language representations. 
Some methods use explicit human actions extracted from mocap videos~\citep{qin2022dexmv,chen2022dextransfer,wang2023mimicplay,kareer2025egomimic,chen2025vividex,qiu2025humanoid} or web videos~\citep{patel2022learning,shaw2023videodex} to guide robot policy training with imitation learning frameworks. 
Instead of using explicit motions, other approaches learn affordances~\citep{bahl2023affordances,mandikal2022dexvip,kannan2023deft,chen2025vidbot}, point trajectories~\citep{bharadhwaj2024track2act,wen2023any,yang2025magma}, or hand-object masks~\citep{singh2025hand} from human videos. Recently, a group of methods have emerged which learn latent actions from human videos in an unsupervised manner and pretrain action model with latent action labels~\citep{yelatent2025,bjorck2025gr00t,chen2024igor,zheng2025flare,bu2025univla}. 
Some recent attempts use extracted 3D hand action labels from egocentric human videos for VLA pretraining~\citep{yang2025egovla,luo2025being,bi2025h}. As mentioned earlier, these primarily involve videos captured in controlled environments with privileged information. In a different vein, some approaches utilize human videos to train video generation models for human-to-robot video transfer~\citep{xiong2021learning,xie2025human2robot}, visual task planning~\citep{bharadhwaj2024gen2act,hu2025video}, or world models~\citep{mendonca2023structured,chen2024igor,jang2025dreamgen}.

\vspace{-4pt}
\paragraph{Temporal Action Segmentation for Videos}
Temporal action segmentation, also known as temporal action detection or localization, is a technique for detecting action windows and classifying them from a long human video. 
Earlier approaches~\citep{wang2014action,shou2016temporal,farha2019ms,zhang2022actionformer,liu2023diffusion}
have focused on predefined action classes. 
Recently, video-input VLMs \citep{chen2024videollm,chen2023videollm} with broad action understanding capabilities are proposed but they still face challenges in action localization accuracy. They do not meet our requirements in our preliminary tests.

%% file: sections/3_method.tex
\section{\!\!Transforming Human Hand Video to VLA Data\!}\label{sec:video2vla}

\begin{figure*}
\vspace{3pt}
	\centering
\includegraphics[width=0.99\textwidth]{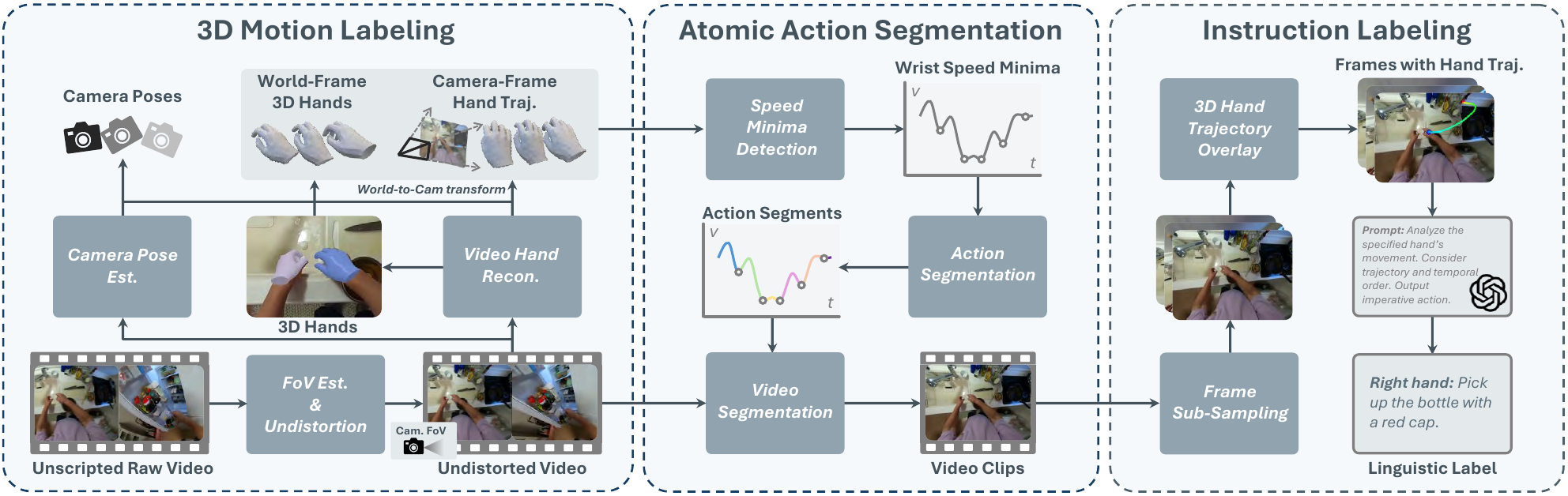}
\vspace{-4pt}
	\captionof{figure}{Overview of our holistic human activity analysis framework, which transforms unscripted real-life human videos into V-L-A episodes of human hands aligned with typical robotic data via three stages: (a) 3D motion labeling, reconstructing metric-scale 3D hand and camera trajectories; (b) atomic action segmentation, dividing videos into atomic-level clips; and (c) instruction labeling, employing GPT to annotate action instructions for each clip.}
	\label{fig:pipeline}
	\vspace{-7pt}
\end{figure*}

Existing robotic manipulation V-L-A data~\citep{zitkovich2023rt,o2024open,bu2025agibot,khazatsky2024droid,fang2024rh20t} typically comprise simple, short-horizon tasks (\emph{e.g.,} \emph{``pick up the sponge on table''}, \emph{``wipe the stove with cloth''}), which can be composed to long-horizon tasks by a high-level planner.
Each data episode comprises a language instruction, a video frame sequence, and frame-aligned 3D action chunks of the end-effector in the robot or camera coordinate system. 
Our approach analyzes an unscripted human video and generates V-L-A data in such format, treating the two human hands as the end-effector. The whole framework comprises three stages and an overview is shown in Fig.~\ref{fig:pipeline}.

\subsection{3D Motion Labeling} 
The first stage of our approach extracts 3D motions from videos, including the motions of two hands and the camera. To achieve this, we first apply a simple algorithm to determine whether the camera is static or moving based on background optical flow. Then we estimate camera intrinsics of the videos by applying DroidCalib~\citep{hagemann2023deep} for moving cameras and MoGe-2~\citep{wang2025moge} and DeepCalib~\citep{bogdan2018deepcalib} for static cameras. The videos with large distortions are then undistorted to conform to the pinhole camera model. Given the intrinsics and undistorted video, we proceed with video hand reconstruction and camera pose tracking. For the former, we employ HaWoR~\citep{zhang2025hawor} to reconstruct per-frame camera-space 3D hands. Each reconstructed hand contains wrist 6D pose and joint angles represented with the MANO~\citep{romero2017embodied} hand parametric model.
To track camera pose for moving cameras, we apply a modified version of MegaSAM~\citep{li2025megasam}, in which we replace the depth estimation model providing depth priors for visual SLAM with MoGe-2~\citep{wang2025moge}. Then we can obtain a sequence of world-space 3D hands by combining the camera-space 3D hands and metric-scale camera poses. Finally, we apply spline smoothing to the world-space hand motions and remove outliers. More details can be found in Appendix~\ref{sec:more_implement_data}.

The world-space 3D hand sequence can be easily transformed into any video frame’s camera space, effectively simulating a static camera as in most robot data. 
Moreover, it facilitates both the subsequent atomic action segmentation and instruction labeling, as will be described later. To enhance efficiency, we chop long videos into overlapping 20-second clips in this stage and recompose their results.

\subsection{Atomic Action Segmentation} 
This stage aims to segment out simple, atomic-level hand action sequences from a long video, in line with the granularity and time windows of robotic V-L-A data. 
This is not a straightforward task and there's no off-the-shelf models that can be applied to this problem reliably. Our solution is inspired by the natural \emph{``beats''} of human hand action in real life. Specifically, during action transitions, human hands typically exhibit speed changes, with minima often indicating switches of action.  This observation has inspired us to leverage the recovered 3D hand motions and design the following algorithm which is simple yet surprisingly effective: \emph{we detect speed minima of the 3D hand wrists in the world space and use them as cutting points.} We smooth the hand trajectory and select points that are local speed minima within a fixed window centered on each point. Segmentation is applied for the left and right hands independently with the other hand's motion ignored. This way, each segment captures a single atomic action of at least one hand.

It is worth noting that this method is highly efficient and requires no additional model inference or pre-annotated text labels, making it particularly effective for the scalable segmentation of hand activity videos. Furthermore, segmenting atomic-level action clips can help reduce the complexity of subsequent instruction captioning, as we discuss in a later section. This strategy may lead to over segmentation for certain actions (\eg, consider a wiping action where a hand moves back and forth), but these actions can be easily merged later after instruction labeling.

\subsection{Instruction Labeling} 
Given the video segments and 3D hand action sequences, we create visualizations and utilize  GPT-4.1~\citep{achiam2023gpt} for action captioning. From each segment, we evenly sample 8 frames and highlight hand trajectories on each frame by  projecting the world-space trajectory of the hand palm from the current frame to the end of the clip (see Fig.~\ref{fig:pipeline} for an example). 
These frames are then fed into GPT, which is prompted to describe the specified hand’s action in imperative form,  taking into account both the content of the frames and the overlaid trajectories. We also instruct GPT to label clips lacking semantically meaningful action as ``N/A''. A detailed description of the prompt design can be found in the Appendix~\ref{sec:more_implement_data}.

We empirically find that providing GPT with atomic-level video clips for captioning is effective in improving annotation accuracy. By contrast, simply splitting the video into fixed-length segments (\eg, 1-second) reduces accuracy, likely because each segment may still contain multiple atomic actions, which increases the difficulty for GPT to reason about the content. Additionally, overlaying hand trajectories on the images is also important for ensuring correct captioning, as evidenced by prior studies incorporating visual markers as supplementary prompts~\citep{yang2023set}.

\subsection{Hand V-L-A Dataset Construction} \label{sec:human_data}
Leveraging the above framework, we construct a large-scale human hand V-L-A dataset by processing egocentric human videos from Ego4D~\citep{grauman2022ego4d}, Epic-Kitchen~\citep{damen2020epic}, EgoExo4D~\citep{grauman2024ego}, and Something-Something-V2 (SSV2)~\citep{goyal2017something}. Note that \emph{the human annotations for actions provided by these datasets are NOT used in this work; instead, we process the raw videos through our framework}. These annotations often do not match the desired task granularity or they lack precise start and end times for actions. Later we'll show in our experiments that training using these annotations results in obvious performance degradation compared to our approach.
Our constructed dataset contains 1M episodes with 26M frames (77\% from Ego4D, 12\% from Epic-Kitchen, 6\% from EgoExo4D, and 5\% from SSV2).
It features diverse hand actions, objects, attributes, and environments, encompassing real-life activities such as cooking, cleaning, construction, repairing, crafting, and painting (Fig.~\ref{fig:teaser}). A more detailed analysis of the dataset will be presented in Sec.~\ref{sec:dataset}.

\section{Dexterous Hand VLA Model}\label{sec:method}

We construct a VLA model $\bm{\pi}$ for dexterous manipulation:\vspace{-2pt}
\begin{equation}
    \bm{\pi}: (\bm{l},\bm{o}_t, \bm{s}_t) \rightarrow (\bm{a}_t,\bm{a}_{t+1},...,\bm{a}_{t+N}), \label{eq:goal}
    \vspace{-2pt}
\end{equation}
which predicts a sequence of future end-effector actions $\bm{a}$ based on the current visual observation $\bm{o}_t$, the robot proprioceptive state $\bm{s}_t$, and a language instruction $\bm{l}$. 

\begin{figure}
\vspace{5pt}
	\centering
\includegraphics[width=0.85\linewidth]{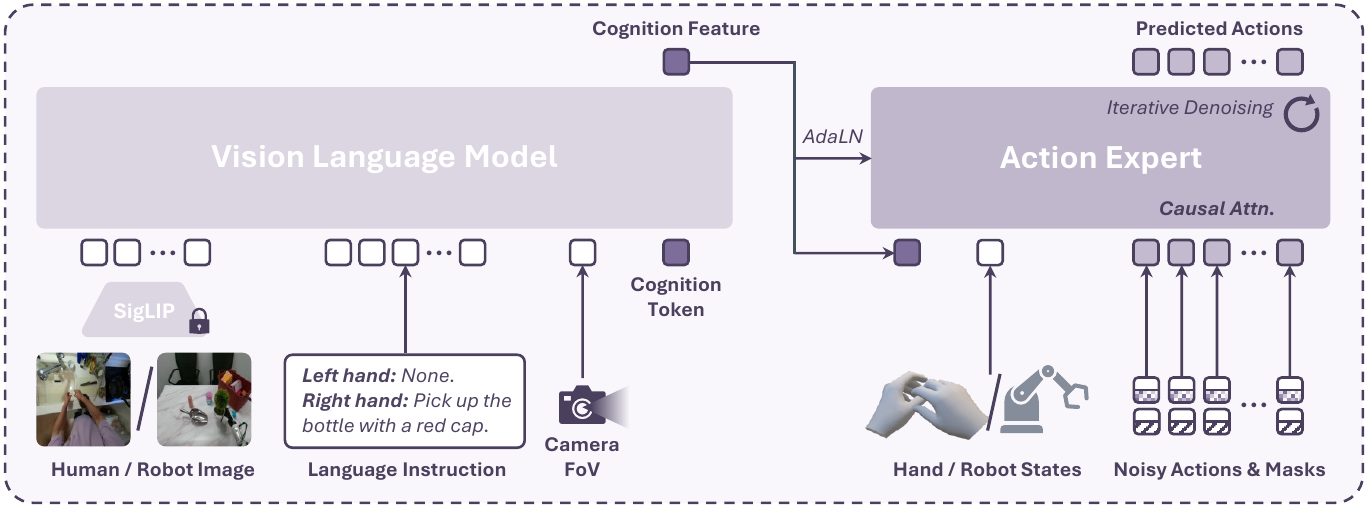}
	\captionof{figure}{Our VLA model architecture. It consists of a VLM backbone and a diffusion action expert. The VLM receives visual and linguistic instructions, as well as the camera FoV, and outputs a cognition feature that guides the action expert for future action chunk prediction. The action expert additionally receives the current state of the end effector and valid action masks for iterative action denoising via causal attention.  }
	\vspace{-6pt}
	\label{fig:model}
	\vspace{-2pt}
\end{figure}

\subsection{VLA Model Design}

\subsubsection{Model Architecture} An overview of our model architecture is presented in Fig.~\ref{fig:model}. Our model consists of a VLM backbone and a diffusion action expert. We use PaliGemma-2~\citep{steiner2024paligemma} as the VLM, which combines a SigLIP~\citep{zhai2023sigmoid} vision encoder with linear projection for alignment and a Gemma-2~\citep{team2024gemma} language model for multi-modal token processing. We use the 3B-parameter model with an input image resolution of $224^2$ as the default setting.
We further incorporate camera FoV information as an extra token to the model to help it better interpret the original image’s aspect ratio and camera intrinsics.  
Following~\citep{li2024cogact}, we append a learnable ``cognition'' token as extra input to the VLM, whose output feature $\bm f^c$ serves as the condition for the action expert.

For the action expert, we apply a Diffusion Transformer (DiT)~\citep{peebles2023scalable} and the DiT-Base model is used by default. The input is a concatenation of the cognition feature $\bm f^c$, the hand state $\bm s_t$, and a noisy action chunk $(\bm a_t^i, \bm a_{t+1}^i, \ldots, \bm a_{t+N}^i)$, where $i$ denotes the denoising step. The hand state includes the wrist translation and rotation in camera space of the current image observation, as well as hand joint angles. A set of action masks, indicating whether each action is valid, is also provided to the model and will be discussed in detail later.
We additionally inject the cognition feature into the DiT using AdaLN~\citep{peebles2023scalable} for enhanced conditioning.
The action expert predicts the added noise for iterative denoising, trained by optimizing an MSE loss: 
\begin{equation}
   \mathcal{L}_{\text{MSE}} = \mathbb{E}_{\bm{\epsilon}\sim\mathcal{N}(0,1),i} ||\boldsymbol{\hat{\epsilon}}^i - \boldsymbol{\epsilon} ||_2, 
\end{equation}
where $\boldsymbol{\hat{\epsilon}}^i$ and $\boldsymbol{\epsilon}$ denote the predicted and ground-truth noise, respectively. The action expert, the VLM, and the cognition token are trained end-to-end, while the vision encoder remains frozen. See Appendix~\ref{sec:more_implement_model} for more implementation details.

\vspace{2pt}
\subsubsection{Hand Action Space} Our model predicts hand actions in the camera coordinate frame of the current observation $\bm o_t$. At time step $t$, the hand action $\bm a_t$ is defined as:
\begin{equation}
    \bm{a}_t = [\Delta \bm t^l, \Delta \bm r^l, \bm \theta_h^l, \Delta \bm t^r, \Delta \bm r^r, \bm \theta_h^r] \in \mathbb{R}^{102}, \label{eq:action}
\end{equation}
where $\Delta \bm t \in \mathbb{R}^3$ and $\Delta \bm r \in \mathbb{R}^3$ are the relative wrist translation and rotation (Euler angles converted from rotation matrices) between the consecutive frames, and $\bm \theta_h \in \mathbb{R}^{15\times3}$ represents the Euler angles of 15 joints in the local frame of the MANO hand model (see Fig.~\ref{fig:xhand} for an illustration). Superscripts $l$ and $r$ indicate the left and right hand, respectively. 

\vspace{2pt}
\subsubsection{Unified Single- and Dual-Hand Action Prediction} Our VLA pretraining data is at the level of single-hand atomic actions, with some episodes containing overlapping dual-hand actions. 
We introduce the following designs to handle different cases in a unified manner.
Specifically, the VLM always receives language instructions in the format of \texttt{Left hand: <left-hand action>. Right hand: <right-hand action>}. 
For a video frame $\bm{o}_t$, left- and right-hand action descriptions are set to either \texttt{None} or the instructions of the corresponding atomic action chuck $\bm{o}_t$ falls within. Meanwhile, the action expert always receives noisy hand actions for both hands. To account for episodes where action labels are available for only one hand, extra action masks (0 or 1), matching the dimensionality of the hand actions, are concatenated with the noisy actions along the feature dimension as input to the action expert (see Fig.~\ref{fig:model}). When a mask value is 0, the corresponding noisy action is set to 0 and excluded from the loss computation.

\vspace{1pt}
\subsubsection{Causal Action Denoising} \label{sec:causal}
Human hands move fast in real life activities, and many of the action clips in our pretraining dataset are as short as 1 second ($\sim$30 frames). Consequently, many prediction chunks of the VLA model go beyond the episode end for a reasonable chunk length (\emph{e.g.,} $N=16$ in our setting). Naively padding with zero actions at the end can be problematic, as many atomic actions occur mid-task and should not conclude with no motion (\emph{e.g.}, a wiping task with a hand moving back and forth). 
To address this issue, we employ  \emph{causal attention} for action denoising, ensuring that the token of each action step only attends to preceding actions. This prevents zero-padded positions from affecting earlier predictions, unlike in the bidirectional attention setting. Furthermore, these padded positions are also excluded from the loss computation with their corresponding action masks set to 0.

\subsection{Pretraining with Human Hand VLA Data}\label{sec:pretrain} 

We first train the VLA model for human hand action prediction using the dataset constructed in Sec.~\ref{sec:human_data}. During training, we apply \emph{trajectory-aware augmentation} to the training images and actions to enhance generalization.
Specifically, input images are randomly cropped and perspective-warped with varying FoV, aspect ratio, and crop center, while keeping the principal point at the image center. The action sequences are transformed accordingly to match the augmented camera parameters. During random cropping, we ensure that the projected hand trajectory from the current frame to episode end remains within the cropped image. Using this strategy, the objects of interaction are also mostly well-contained. We also apply random image flipping and make corresponding adjustments to hand actions and language instructions. Random color jittering is further applied when the text description does not contain explicit color cues.

\subsection{Fine-tuning for Robotic Dexterous Manipulation}\label{sec:finetune}
After pretraining, the model can be fine-tuned on robot data for deployment. 
We consider the human hand action space as a superset of that of the robot hand and align the robot's action space with the human hand’s as defined in Eq.~\eqref{eq:action}. Specifically, robot end-effector 6D poses in camera coordinates are used to compute $\Delta \bm t$ and $\Delta \bm r$. 
For joint angles, a simple mapping strategy is applied: each joint of robot hand is mapped to its closest human joint in topology, and the corresponding dimension in human action $\bm{\theta}_h$ is used for fine-tuning (an example is shown in Fig.~\ref{fig:xhand}). Unmapped dimensions in $\bm{\theta}_h$ are zero-padded in the action mask. In addition, we supervise the model with direct future execution commands for the hand joints, instead of using action labels derived from the recorded robot states. This approach produces more plausible hand motions during hand-object interactions. The language instruction format is kept consistent with those used during pretraining.

\emph{A Remark.} Action space mapping between human hand and dexterous robot hand have been actively studied in the past~\citep{handa2020dexpilot,wang2024dexcap,qiu2025humanoid}. In this work we do not perform direct pose transfer (as done in teleoporation) and our fine-tuning can help mitigate the action space differences. Other strategies can also be employed and we leave it as our future work.

%% file: sections/4_experiment.tex
\section{Experiments}

{\emph{Training Details}}.~For pretraining, we first warm up the action expert, the mapping layers of the cognition token, and the MLP projecting FoV for 5K steps. Then we jointly fine-tune the VLM backbone and action expert for 80K steps. The learning rates are 1e-4 and 1e-5 for the action expert and VLM, respectively, with a batch size of 512. The pretraining stage takes 2 days on 8 NVIDIA H100 GPUs. 
For fine-tuning on real robot data, we optimize the model for 20K steps with a batch size of 256 and a learning rate of 1e-5, which takes 8 hours with 8 NVIDIA H100 GPUs. More details can be found in Appendix~\ref{sec:more_implement_train}.

\subsection{Pretraining Data Analysis} \label{sec:dataset}
An overview of our pretraining data is shown in Fig.~\ref{fig:teaser}. We visualize the most frequent words in the language instructions using word clouds, and showcase randomly-sampled task environments. More examples of the dataset can be found in the Appendix~\ref{sec:more_results_data}. To further investigate data diversity, we conduct a detailed analysis of the visual observations and language instructions in the dataset, as described below.

\subsubsection{Visual Diversity} \label{sec:visual_diversity} The diversity of visual observations and their coverage of natural scenes are crucial for enhancing the model’s generalization ability in real-world scenarios. To quantify the diversity and coverage of our dataset, we use the OpenImages~\citep{kuznetsova2020open} dataset as a reference and compute the similarity between our dataset and it, as OpenImages spans a broad spectrum of real-world scenes and is known to be highly diverse~\citep{xing2025shortcut}. 
Specifically, we randomly sample 8K images from OpenImages as queries and extract their features using the DINOv2~\citep{oquab2024dinov2} encoder. For each query feature, we compute its maximum cosine similarity to our dataset, where the target features are extracted from the first frame of each episode in the dataset. We use the average of the maximum cosine similarities for all query features as a measure of dataset diversity. Higher similarity values indicate that the dataset covers a larger portion of real-world scenes represented in OpenImages.

\begin{figure*}[t!]
    \centering

    \begin{subfigure}[t]{0.4\linewidth}
        \centering
        \includegraphics[height=5cm]{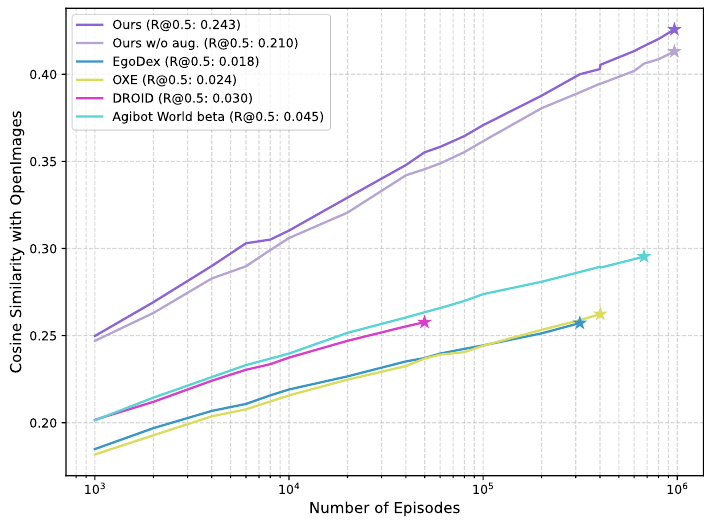}
    \captionsetup{margin={0cm,-0.5cm}}
        \caption{}
        \label{fig:data_analysis_image_1}
    \end{subfigure}
    \hfill
    \begin{subfigure}[t]{0.58\linewidth}
        \centering
        \includegraphics[height=5cm]{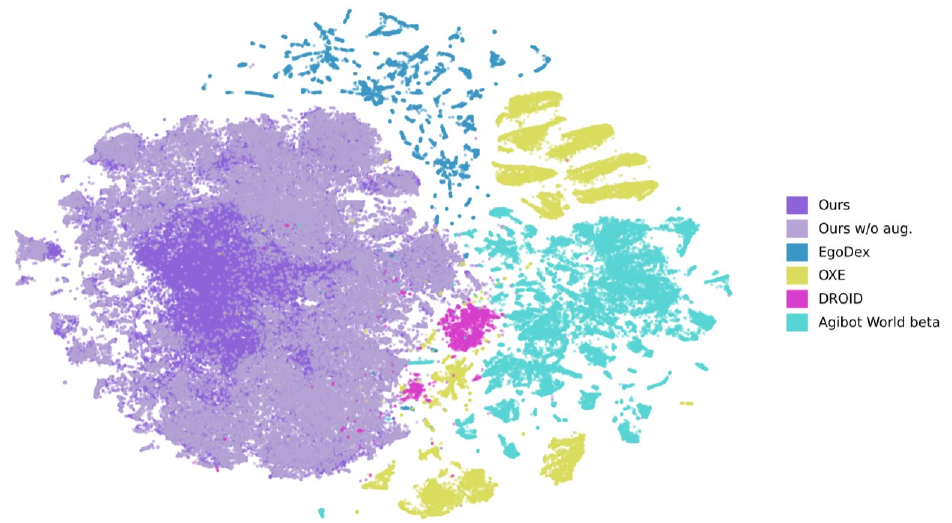}
    \captionsetup{margin={0cm,1cm}}
        \caption{}
        \label{fig:data_analysis_image_2}
    \end{subfigure}
    \vspace{-4pt}
    \caption{Visual diversity across VLA datasets. (a) Image feature similarity with OpenImages~\citep{kuznetsova2020open} as the number of episodes varies. $\star$ marks the full dataset's similarity. (b) t-SNE visualization of image features.}
    \label{fig:data_analysis_image}
\end{figure*}

\begin{figure*}[t!]
    \centering

    \begin{subfigure}[t]{0.32\linewidth}
        \centering
        \includegraphics[height=5cm]{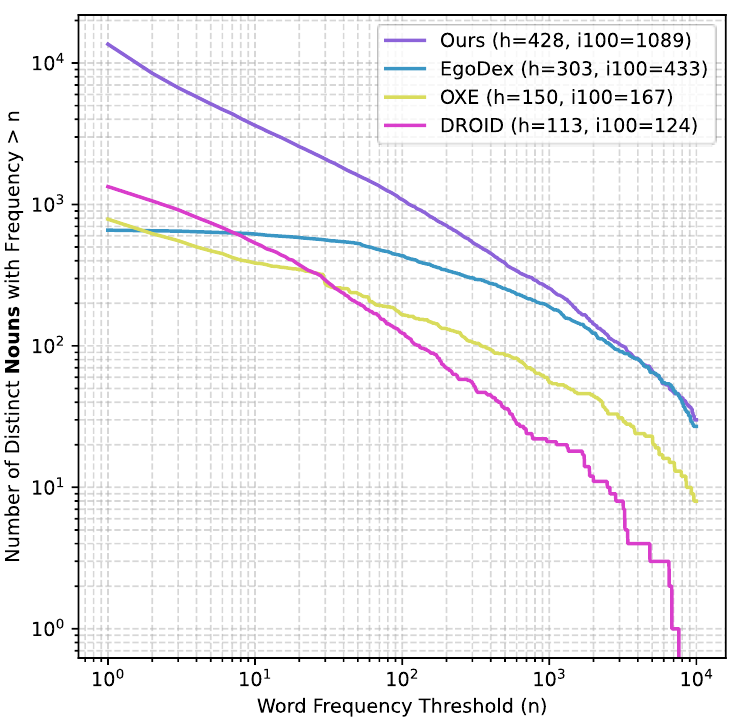}
    \captionsetup{margin={0cm,-0.5cm}}
        \caption{Nouns}
        \label{fig:data_analysis_text_1}
    \end{subfigure}
    \hfill
    \begin{subfigure}[t]{0.32\linewidth}
        \centering
        \includegraphics[height=5cm]{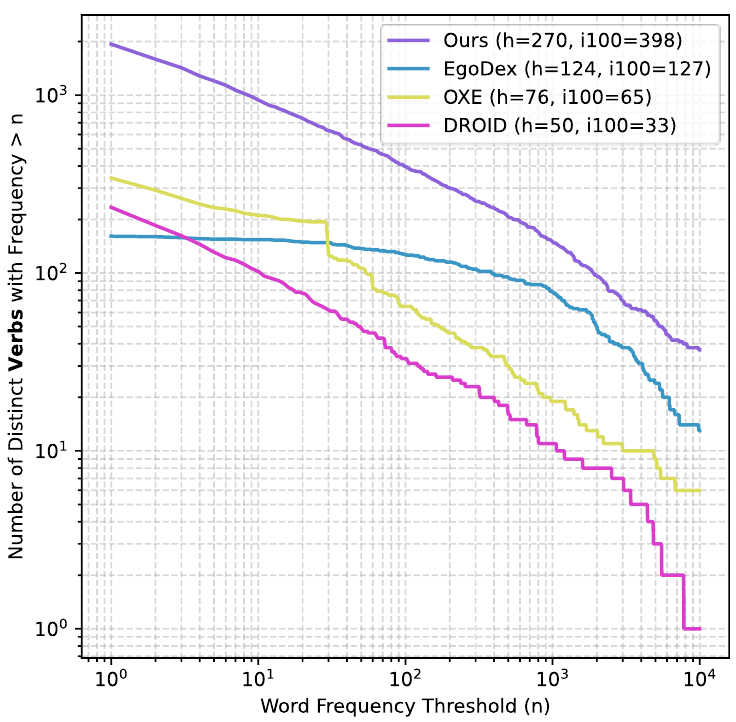}
    \captionsetup{margin={0cm,-0.5cm}}
        \caption{Verbs}
        \label{fig:data_analysis_text_2}
    \end{subfigure}
    \hfill
    \begin{subfigure}[t]{0.32\linewidth}
        \centering
        \includegraphics[height=5cm]{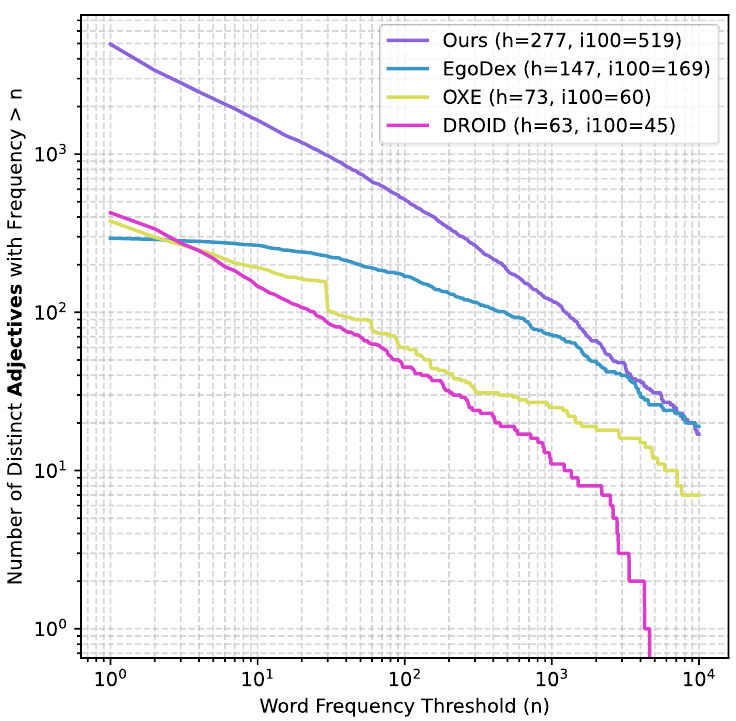}
    \captionsetup{margin={0cm,-0.5cm}}
        \caption{Adjectives}
        \label{fig:data_analysis_text_3}
    \end{subfigure}
    \vspace{-4pt}
    \caption{Language instruction statistics across different VLA datasets.}
    \label{fig:data_analysis_text}
\end{figure*}

Figure~\ref{fig:data_analysis_image_1} presents the similarity curve as a function of the number of episodes, where we randomly sample varying numbers of episodes from the dataset and compute feature similarity as described above. We also report $\text{R@0.5}$, \ie, the fraction of OpenImages queries with a maximum similarity above 0.5 to the target dataset features. We compare our dataset with existing VLA datasets, including EgoDex~\citep{hoque2025egodex}, a human-hand VLA dataset of over 300K episodes collected in lab environments, and widely-used robotic VLA datasets: Open X-Embodiment (OXE)\footnote{We use a subset comprising approximately 400K episodes widely used for VLA pretraining, following~\citep{li2024cogact,kim2024openvla}.}~\citep{o2024open}, DROID~\citep{khazatsky2024droid}, and AgiBot World beta~\citep{bu2025agibot}. As shown in the figure, our dataset exhibits higher similarity to the OpenImages dataset, indicating greater diversity and broader coverage of real-world scenes. Even when sampling a small subset of our dataset (10K episodes), its diversity already surpasses that of the other datasets. In addition, the diversity of visual observations can be further increased by leveraging the augmentation strategies described in Sec.~\ref{sec:pretrain}. Moreover, our similarity increases more rapidly with the number of episodes (\ie, with a steeper slope), indicating a more uniform coverage of real-world scenes, in contrast to the fragmented distribution observed in OXE~\citep{xing2025shortcut}. The t-SNE visualization of image features in Fig.~\ref{fig:data_analysis_image_2} also aligns with the observations.

\subsubsection{Instruction Diversity} 
The diversity of language instructions is also important for the model to perform a wide range of tasks. To fairly compare datasets with varying instruction formats, we employ GPT-4.1 to extract nouns, verbs, and adjectives from each instruction and analyze their distributions separately. Figure~\ref{fig:data_analysis_text} illustrates the relationship between the number of distinct words and their frequency of occurrence. A dataset with high diversity should contain a large number of distinct words, each appearing with sufficient frequency. Accordingly, curves positioned closer to the upper-right corner indicate a higher degree of instruction diversity. As illustrated, our dataset demonstrates a significantly higher degree of diversity than existing human and robot VLA datasets. Additionally, we compute the h-index and i100-index for the words, where the h-index represents the largest number $h$ such that at least $h$ words appear at least $h$ times, and the i100-index counts the number of words appearing at least 100 times. Our dataset achieves higher values on both metrics compared to the other datasets.

\subsection{Human Hand Action Prediction}\label{humanhandprediction}

In this section, we evaluate the performance of our pretrained VLA model for human hand action prediction in \emph{unseen environments}. We examine how action prediction performance is influenced by key factors such as dataset composition, model architecture, training strategies, data construction strategies, and dataset scale. We first introduce the benchmark used for evaluating hand action prediction, followed by a systematic analysis of how each factor impacts performance.

\begin{figure*}[t!]
    \centering

    \begin{subfigure}[t]{0.56\linewidth}
        \centering
        \includegraphics[height=5.2cm]{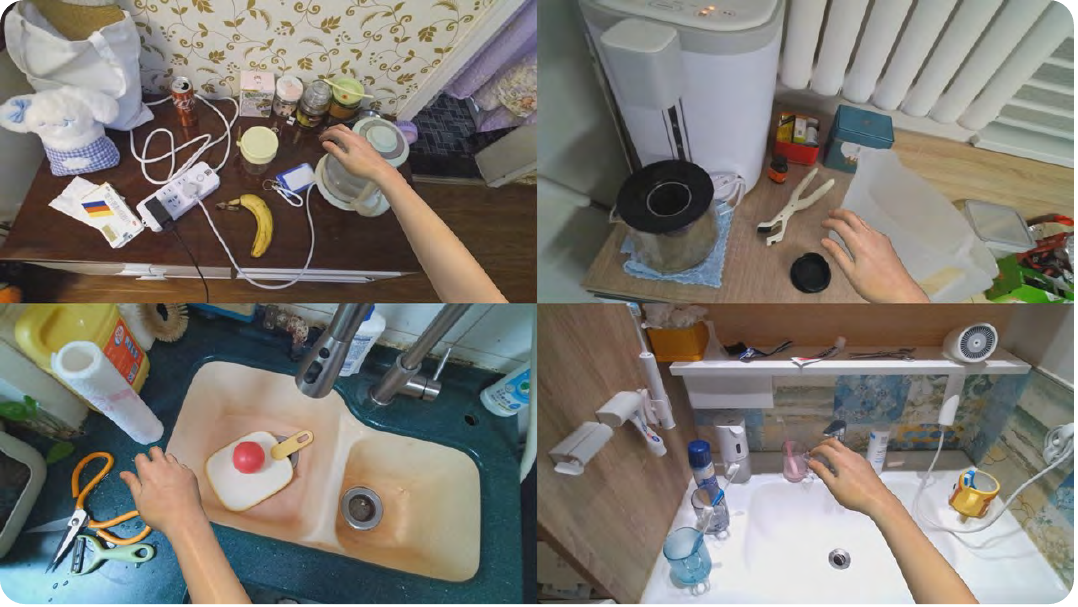}
        \caption{Grasping}
        \label{fig:benchmark_quat}
    \end{subfigure}
    \hfill
    \begin{subfigure}[t]{0.42\linewidth}
        \centering
        \includegraphics[height=5.2cm]{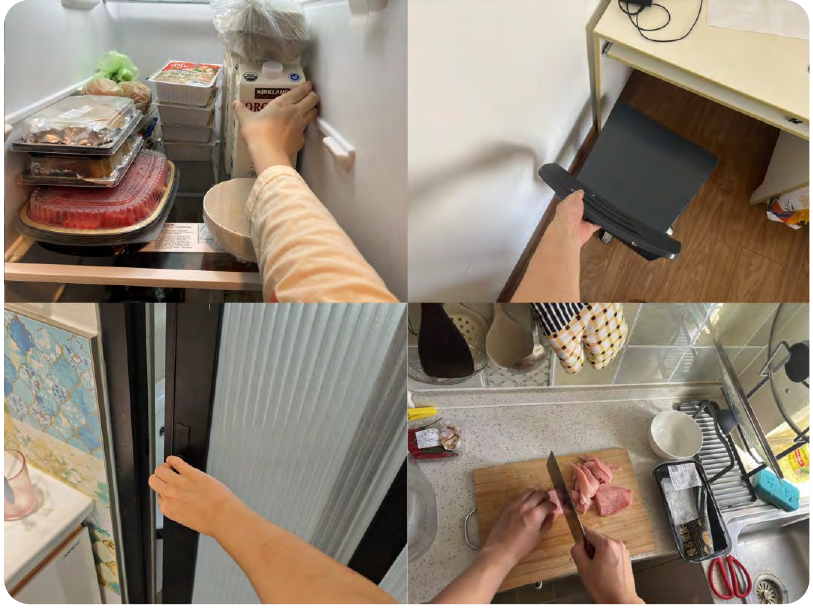}
        \caption{General actions}
        \label{fig:benchmark_user}
    \end{subfigure}
    \vspace{-4pt}
    \caption{Examples of environments used in hand action prediction evaluation.}
    \label{fig:benchmark}

\end{figure*}

\subsubsection{Benchmark}\label{sec:hand_benchmark} We construct a benchmark under \emph{unseen real-life environments}, consisting of two task types defined below:

\paragraph{Grasping} We instruct the model to grasp objects in the scene. We capture RGB-D images from 47 unseen environments using Azure Kinect and annotate 396 objects with captions and segmented 3D point clouds. Synthetic human hands are rendered onto the images at distances suitable for object grasping with a single action chunk (see Fig.~\ref{fig:benchmark_quat} for examples). We compute the minimum distance between predicted finger trajectories and target object points (\ie, $d_{\text{hand-obj}}$) to evaluate movement plausibility.

\paragraph{General Action} For more general hand actions, quantitative metrics may not adequately capture the plausibility or correctness of the predicted movements. To address this, we design a user study to evaluate hand movements before and after contact across 117 unseen real-life environments captured with mobile phones (see Fig.~\ref{fig:teaser} and Fig.~\ref{fig:benchmark_user}). For each scene, we prompt the model with annotated instructions and render predicted hand actions onto video frames. We ask 23 participants to rank the top-3 actions for 30 randomly selected scenes from the 117 environments. These actions will be assigned 3, 2, and 1 scores while all others receive 0. We then report the average scores across participants to assess the plausibility of the predicted actions.

More details of the benchmark can be found in Appendix~\ref{sec:more_eval}.

\begin{table}[t!]  
	\centering  
    \small
    \vspace{3pt}
    \caption{Evaluation and ablation study of hand action prediction for the pretrained model. Note that Being-H0~\citep{luo2025being} is a concurrent work to ours. See text for details.}  
    \vspace{-4pt}
    \renewcommand{\arraystretch}{0.9}
    \setlength{\tabcolsep}{0.5pt}
	\begin{tabular}{l |c |c}  
        \toprule
        \multirow{2}{*}{Method} 
            & Grasp & \quad General action \quad \\ 
             \cmidrule{2-2} \cmidrule{3-3}  
        & \quad Avg./med. $d_{\text{hand-obj}}$ (cm) $\downarrow$ \quad & \quad User Score $\uparrow$ \quad \\
		 \midrule  
      Initial position  &  20.0 / 20.0  & --\\
      \midrule
    Being-H0 (8B)  & 19.1 / 18.4 & 0.15 \\
    \midrule
    {\scriptsize \emph{Ablations}} & & \\
      ~~~Lab data (EgoDex)\quad\quad & 17.6 / 18.3 & -- \\
      ~~~Human annotation\quad\quad & 14.1 / 14.1 & 0.96\\
      ~~~No augmentation \quad\quad& 11.6 / 10.7 & 1.43\\
      ~~~Bidirectional attention \quad\quad  & \hspace{0.5em}9.3 / \hspace{0.5em}7.2 & 1.69  \\
      ~~~\textbf{\textit{Ours}}& \textbf{\hspace{0.3em}8.8 / 6.2} & \textbf{1.91}\\
		\bottomrule
	\end{tabular}  
    \vspace{-3pt}
	\label{tab:pretrain}  
\end{table}

\begin{figure}[t!]
    \centering
    \begin{minipage}[t]{0.54\textwidth}
        \centering
        \small
        \vspace{0pt} 
        \captionof{table}{Ablation study on hand action prediction using different episode construction strategies. $\dag$: Pretrained using a subset of 350K episodes for efficiency. }
        \renewcommand{\arraystretch}{0.9}
        \setlength{\tabcolsep}{0.5pt}
        \begin{tabular}{l |c }  
            \toprule
            \multirow{2}{*}{Method} 
                & Grasp  \\ 
                \cmidrule{2-2}
            & \quad Avg./med. $d_{\text{hand-obj}}$ (cm) $\downarrow$  \\
            \midrule  
            Initial position  &  20.0 / 20.0  \\ 
            \midrule
    {\scriptsize \emph{Ablations}} & \\
            ~~~Fixed-interval segmentation \quad & 10.5 / \hspace{0.5em}8.8 \\
            ~~~No trajectory overlay \quad  & 11.7 / 10.7   \\
            ~~~\textbf{\textit{Ours}$^\dag$}& \textbf{\hspace{0.5em}9.9 / 8.1} \\
            \bottomrule
        \end{tabular}
        \label{tab:episode}
    \end{minipage}%
    \hspace{0.05\textwidth}
    \begin{minipage}[t]{0.38\textwidth}
        \centering
        \caption{Data scaling behavior on the grasping task. The circle size indicates the \emph{visual diversity} of the data.}
        \vspace{-2pt}
        \includegraphics[width=\linewidth]{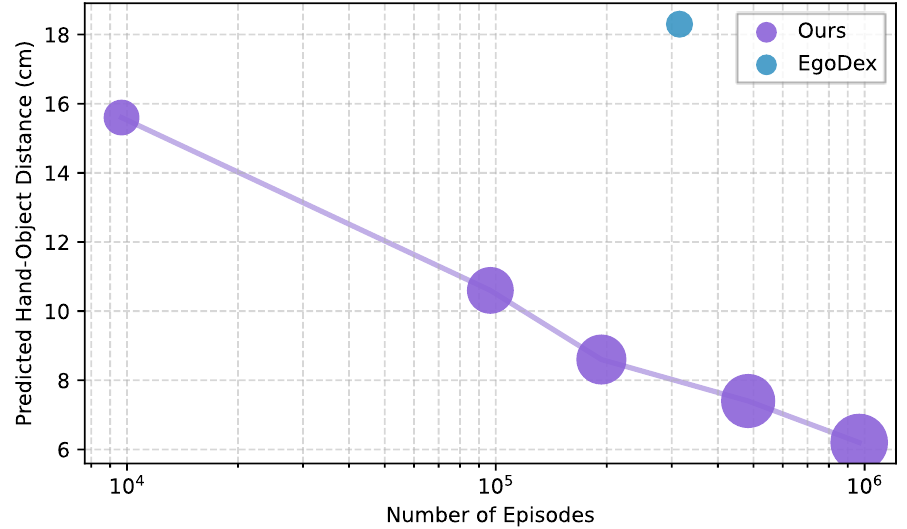} 
        \label{fig:data_scale}
    \end{minipage}
    \vspace{-10pt}
\end{figure}

\subsubsection{Performance Analysis}\label{sec:hand_prediction_result}

\paragraph{Comparison of Pretraining Data} We first compare the performance of models trained with different pretraining datasets to validate the effectiveness of our constructed data. We compared with several baselines including \emph{a) Lab data}, which replaces our VLA data with the EgoDex dataset captured in lab environments; \emph{b) Human annotation}, which uses human annotations in the original human video datasets described in Sec.~\ref{sec:human_data} for constructing VLA episodes (as mentioned previously, these annotations often do not match the desired task granularity or there's no precise start and end times); \emph{c) Being-H0}~\citep{luo2025being}, a recent hand VLA model pretrained on a large collection of scripted, laboratory human video data (as a reference). 

Table~\ref{tab:pretrain} reports the quantitative results of different configurations on our constructed benchmark. For \emph{grasping}, we include the initial hand–object distance as a reference. 
For \emph{general action}, the \emph{Lab data} baseline is not included because the model only predicts keypoints using labels provided by EgoDex.
As shown, our method consistently outperforms all baselines. Compared to models trained on EgoDex data, ours exhibit much stronger generalizability. Being-H0, though pretrained on multiple lab datasets, still shows limited performance. Moreover, training with the original human annotations also underperforms, as its temporal or granularity misalignment between text and actions weakens instruction following. Figure~\ref{fig:teaser} further presents visual results of our model’s hand action predictions in these unseen environments, demonstrating its strong generalization to diverse scenarios. More visualization results can be found in Appendix~\ref{sec:more_results_hand}.

\vspace{-5pt}
\paragraph{Influence of Model Design and Training Strategy} We evaluate the efficacy of model and training designs by comparing with two alternatives: \emph{a) No augmentation}, which discards the trajectory-aware data augmentation during training; \emph{b) Bidirectional attention}, which uses bidirectional attention for action denoising in the diffusion action expert. 

As shown in Tab.~\ref{tab:pretrain}, removing data augmentation during training largely reduces performance. This observation is consistent with the findings in Fig.~\ref{fig:data_analysis_image}, where data augmentation is shown to play a crucial role in enhancing the diversity of visual observations, which is essential for improving the model’s generalization ability in unseen environments. In addition, replacing causal attention with bidirectional attention also leads to a performance drop, highlighting the importance of this technique which better aligns with the characteristics of our pretraining data.

\vspace{-5pt}
\paragraph{Influence of Episode Construction Strategy} Our framework segments unscripted human videos into atomic action episodes and generates their language instructions by leveraging \emph{reconstructed 3D hand trajectories}. We compare our method with two baselines that omit the use of 3D hand trajectory guidance during episode construction: \emph{a) Fixed-interval segmentation}, which segments raw videos into non-overlapping clips of 1 second instead of using speed minima as cutting points, and feeds the resulting clips to GPT for captioning; \emph{b) No trajectory overlay}, which prompts GPT for action captioning without overlaying the projected 3D hand trajectory onto sampled frames. To improve efficiency, this ablation study is conducted on a subset of 350K episodes from Ego4D.

Table~\ref{tab:episode} demonstrates the quantitative results on grasping tasks. Using fixed-interval segmentation for constructing VLA episodes during pretraining results in degraded performance, as this approach can include multiple actions within a single clip, thereby increasing the difficulty for GPT to correctly interpret and align the corresponding action captions. Similarly, removing the hand trajectory overlay during GPT captioning also results in a noticeable performance drop. This highlights the importance of leveraging 3D hand trajectories as guidance in constructing VLA episodes.

\vspace{-6pt}
\paragraph{Data Scaling Behavior} We further investigate how the scale of training data influences hand action prediction performance. We compare the model trained on the full dataset with those trained on sub-sampled datasets at different ratios: 50\%, 20\%, 10\%, and 1\%. 

Figure~\ref{fig:data_scale} presents the trend of performance on the grasping task with respect to the scale of the training data. The size of the circles indicate visual diversity of the training data, as defined in Sec.~\ref{sec:visual_diversity}. We also plot the performance of model trained with EgoDex as a reference. As shown, the predicted hand-object distance steadily decreases with increasing data scale, following an approximately linear trend on the log scale. In addition, although EgoDex contains more episodes than our 20\%, 10\%, and 1\% subsets, its performance still lags significantly behind the models trained on our data. We attribute this mainly to its lower data diversity, which limits its generalization ability in unseen scenarios.

\subsection{Real-World Robot Dexterous Manipulation}\label{sec:exp_robot}

In this section, we evaluate the performance of our VLA model fine-tuned on a small set of real robot trajectories for dexterous manipulation tasks. We begin by describing the hardware system and the tasks defined for real-robot evaluation, followed by a detailed analysis of our model's performance and comparison with prior methods.

\subsubsection{Robot Setup} 
\begin{figure*}[t!]
    \centering
    \begin{subfigure}[t]{0.36\linewidth}
        \centering
        \includegraphics[height=4cm]{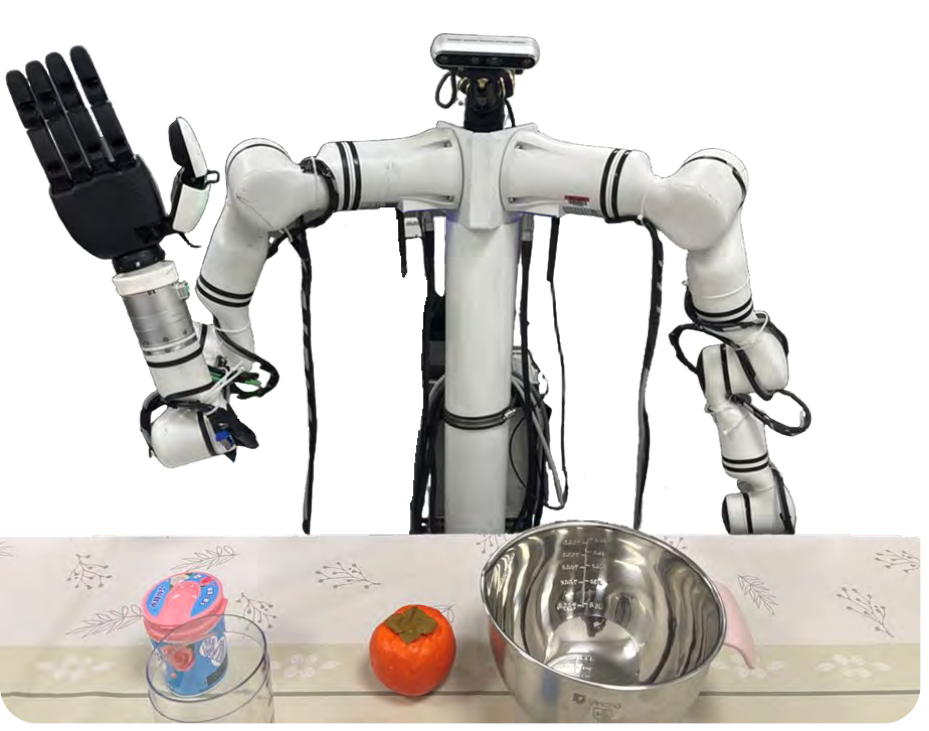}
        \caption{}
        \label{fig:robot_setup}
    \end{subfigure}
    \hfill
    \begin{subfigure}[t]{0.36\linewidth}
        \centering
        \includegraphics[height=4cm]{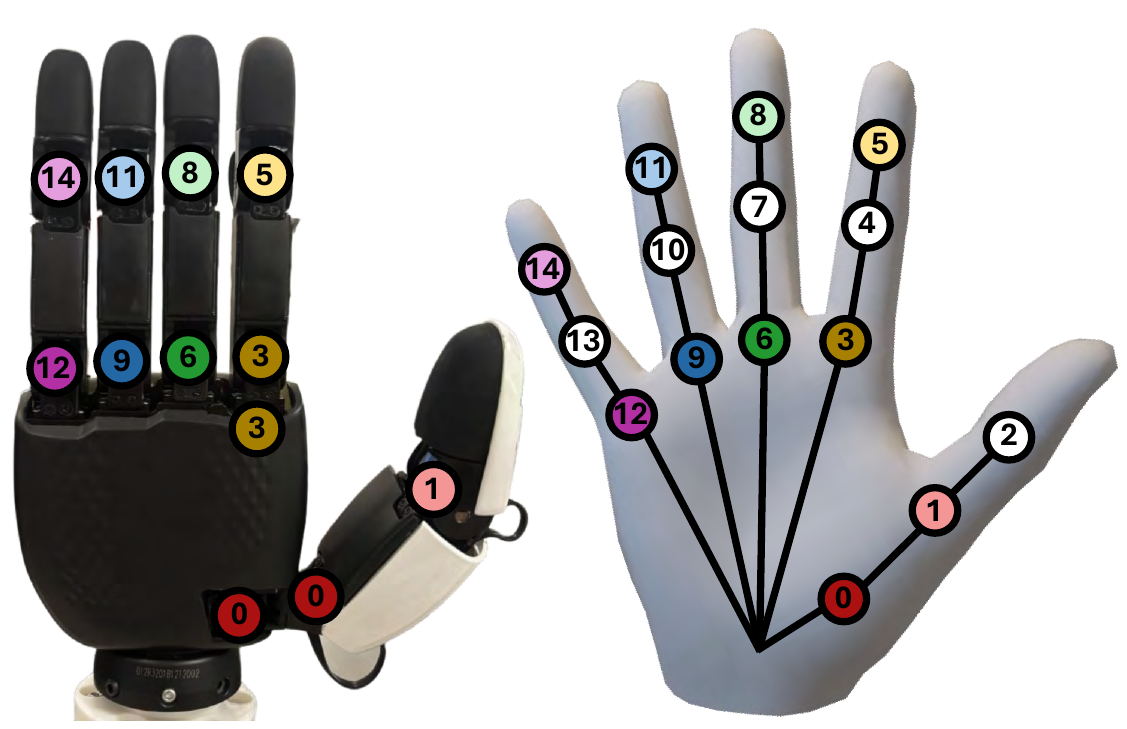}
        \caption{}
        \label{fig:xhand}
    \end{subfigure}
    \hfill
    \begin{subfigure}[t]{0.26\linewidth}
        \centering
        \includegraphics[height=4cm]{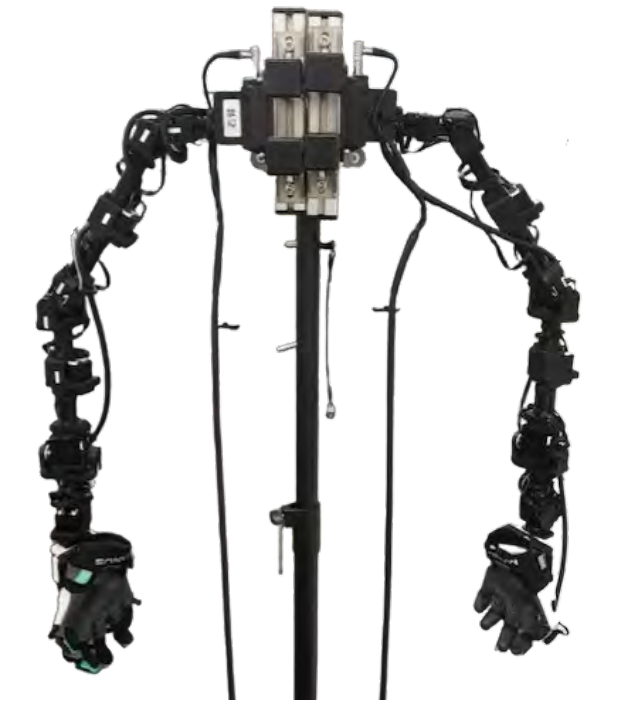}
        \caption{}
        \label{fig:teleop}
    \end{subfigure}
    \vspace{-4pt}
    \caption{(a) Robot setup in our experiments. (b) Mapping between XHand and MANO, where joints sharing the same index indicate correspondence; white color denotes joints without counterparts. (c) Teleoperation hardware system used for robot data collection.}
    \label{fig:robot}
\end{figure*}

We use a Realman\footnote{https://www.realman-robotics.com/rm75-b.html}
 robot equipped with 12-DoF XHand\footnote{https://www.robotera.com/en/goods/2.html}
 dexterous hands and a RealSense head camera, as shown in Fig.~\ref{fig:robot_setup}. The robot is placed in a tabletop environment. The joint mapping between the XHand and a human hand for fine-tuning is illustrated in Fig.~\ref{fig:xhand}. For real-robot data collection, we employ the teleoperation system shown in Fig.~\ref{fig:teleop}. It consists of two teleoperation arms matching the Realman arms’ topology for controlling end-effector 6D poses, and a pair of MANUS\footnote{https://docs.manus-meta.com/latest/Products/} teleoperation gloves at the arm’s end for controlling the dexterous hand joint angles. See Appendix~\ref{sec:more_implement_robot} for more details of the teleoperation system. 

\begin{figure*}[t!]
    \centering
    \begin{subfigure}[t]{0.49\linewidth}
        \centering
        \includegraphics[height=5cm]{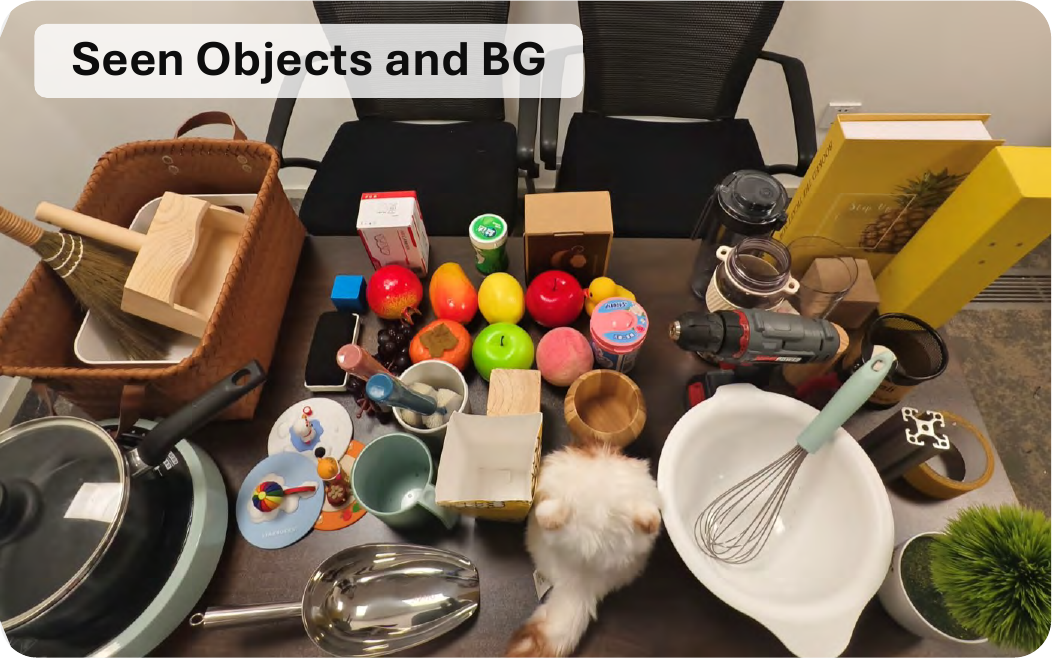}
        \caption{}
        \label{fig:robot_seen}
    \end{subfigure}
    \hfill
    \begin{subfigure}[t]{0.49\linewidth}
        \centering
        \includegraphics[height=5cm]{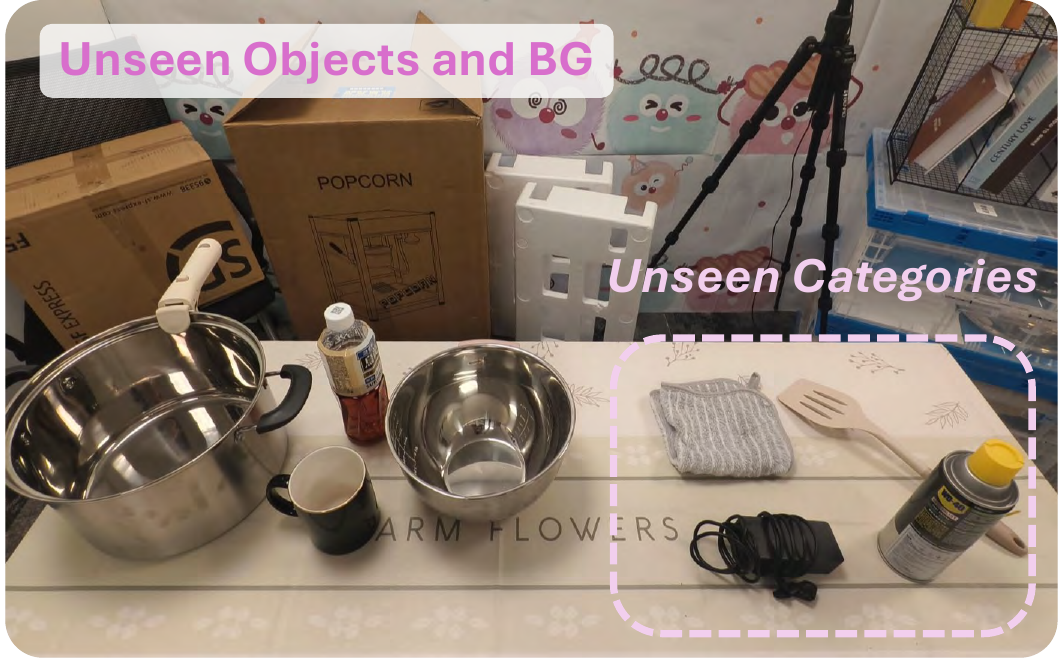}
        \caption{}
        \label{fig:robot_unseen}
    \end{subfigure}
    \vspace{-4pt}
    \caption{Fine-tuning objects/environment and unseen objects/background for real robot evaluation.}
    \label{fig:robot_env}
\end{figure*}

\vspace{2pt}
\subsubsection{Task Designs} We collected 1.2K teleoperated trajectories for four tasks: \emph{i) General pick \& place} -- moving an object into a box with 3–4 random distractors; \emph{ii) Functional grasping} -- grasping an object at a functional location (\eg, handle); \emph{iii) Pouring} -- picking up a bottle, pouring its contents into another container, and placing it back on the table; \emph{iv) Sweeping} -- picking up a broom from a basket, sweeping trash into a dustpan, and returning the broom. Examples of the four tasks are illustrated in Fig.~\ref{fig:teaser}. For evaluation, we perform the above tasks in both \emph{seen} and \emph{unseen} settings (see Fig.~\ref{fig:robot_env} for seen and unseen objects and backgrounds):

\paragraph{Seen} Objects and backgrounds were observed during fine-tuning; randomized positions and distractors are added during evaluation.

\paragraph{Unseen} Novel objects and backgrounds for evaluation, with two additional settings: \emph{Unseen Objects}, where the objects are new but other objects of the same categories were seen in fine-tuning; and \emph{Unseen Categories}, where the objects belong to categories not encountered before.

\vspace{2pt}
\subsubsection{Results and Comparisons} Some representative execution results of our method are presented in Fig.~\ref{fig:teaser}, more are presented in Appendix~\ref{sec:more_results_robot}. For quantitative evaluation, we examine the model’s performance in terms of task success rate and compare with prior methods. We also analyze the effect of different pretraining data and action representations, the data scaling behavior, and the relationship between robot performance and the performance of pretraining human-hand action prediction. All baseline methods are fine-tuned using the same robot data collected in our study for a fair comparison.

 \begin{table*}[t]
\vspace{3pt}
\centering
\small
\renewcommand{\arraystretch}{0.9}
\caption{Success rates on seen real-world robot dexterous manipulation tasks (in \%) .}
\vspace{-6pt}
\begin{tabular}{lccccc}
\toprule
\multirow{3}{*}{Method} 
& \multicolumn{4}{c}{Seen Object} 
& \multirow{3}{*}{\!Average\!}
 \\
\cmidrule(lr){2-5}
& \!Pick \& place\! & \!\!\!Functional grasp\!\!\! & Pour & Sweep  \\
& (40 trials) & (24 trials) & (8 trials) & (8 trials) \\
\midrule
VPP                 & 57.5 & 29.2 & 12.5     & \hspace{0.5em}0.0 & 24.8  \\
$\pi_0$               & 37.5 & 25.0  &  \textbf{75.0}   &  50.0 & 46.9\\
\midrule
No VLA pretrain            & 32.5 & 33.3& 12.5     &  50.0 &32.1  \\
Latent action pretrain\!\!\!  & 42.5 & 41.7  & 37.5     & \textbf{62.5} & 46.0\\
OXE pretrain         & 40.0 & 37.5  & 62.5     &  25.0  & 41.3 \\

\textbf{\textit{Ours}}  & \textbf{80.0} & \textbf{66.7} & \textbf{75.0}     & \textbf{62.5} & \textbf{71.0}  \\
\bottomrule
\end{tabular}
\label{tab:robot}  
\vspace{-2pt}
\end{table*}

\begin{table*}[t]
\vspace{3pt}
\centering
\small
\renewcommand{\arraystretch}{0.9}
\caption{Success rates on unseen real-world robot dexterous manipulation tasks (in \%) .}
\vspace{-6pt}
\begin{tabular}{lcccccc}
\toprule
\multirow{3}{*}{Method} 
& \multicolumn{3}{c}{Unseen Object \& Background} 
& \multicolumn{1}{c}{Unseen Category \& Background} 
& \multirow{3}{*}{\!Average\!} \\
\cmidrule(lr){2-4} \cmidrule(lr){5-5} 
& \!Pick \& place\! & \!\!\!Functional grasp\!\!\! & Pour & \!Pick \& place\! \\
& (16 trials) & (16 trials) & (8 trials) & (24 trials) \\
\midrule
VPP                 & 12.5 & \hspace{0.5em}0.0& \hspace{0.5em}0.0& \hspace{0.5em}8.3 & \hspace{0.5em}5.2 \\
$\pi_0$               &\hspace{0.5em}0.0 &\hspace{0.5em}6.2&25.0& 33.3&16.1 \\
\midrule
No VLA pretrain            &31.2 & \hspace{0.5em}0.0&\hspace{0.5em}0.0&12.5 &10.9  \\
Latent action pretrain\!\!\!  &  \hspace{0.5em}0.0& \hspace{0.5em}0.0& \hspace{0.5em}0.0 & \hspace{0.5em}0.0& \hspace{0.5em}0.0 \\
OXE pretrain         &12.5 &\hspace{0.5em}6.3 &\hspace{0.5em}0.0 & 12.5&\hspace{0.5em}7.8\\

\textbf{\textit{Ours}}  & \textbf{68.8} & \textbf{68.8}&\textbf{50.0}& \textbf{70.8}&\textbf{64.6} \\
\bottomrule
\end{tabular}
\label{tab:robot_unseen}  
\vspace{-4pt}
\end{table*}

\vspace{-3pt}
\paragraph{Comparison with Prior Art} We compare our method with two representative works: \emph{a) VPP}~\citep{hu2025video}, a recent dexterous hand manipulation model leveraging diffusion-based video generation pretraining~\citep{blattmann2023stable}; and \emph{b) $\pi_0$}~\citep{black2410pi0}, a VLA model pretrained on extensive robot data covering a wide range of embodiments.

Table~\ref{tab:robot} and~\ref{tab:robot_unseen} compare the performance of different methods, where our approach achieves significantly higher success rates than VPP and $\pi_0$, especially on unseen tasks. In our tests, the VPP model lags behind LLM-based VLA models in instruction following and unseen object recognition; its implicit supervision through video generation pretraining seems to transfer poorly to real manipulation tasks. While $\pi_0$ is pretrained on large-scale robot data, its knowledge primarily targets gripper-based robots and does not transfer effectively to dexterous hands. Our model demonstrates robust generalization to unseen objects and environmental changes and even for objects from unseen categories, highlighting the effectiveness of leveraging human activity data for generalizable VLA learning.

\vspace{-3pt}
\paragraph{Comparison of Pretraining Data} To investigate the impact of different pretraining data on robot performance, we compare our method with several baselines: \emph{a) No VLA pretrain}, which is directly fine-tuned from the base VLM model (initialized with PaliGemma-2 VLM) and a randomly-initialized action expert without human data pretraining; \emph{b) OXE pretrain}, which uses Open X-Embodiment data instead of our human VLA data for pretraining; and \emph{c) EgoDex pretrain}, which uses EgoDex data under lab environment for pretraining. 

\begin{figure*}[t!]
    \centering
    \begin{subfigure}[t]{0.3\linewidth}
        \centering
        \includegraphics[height=5cm]{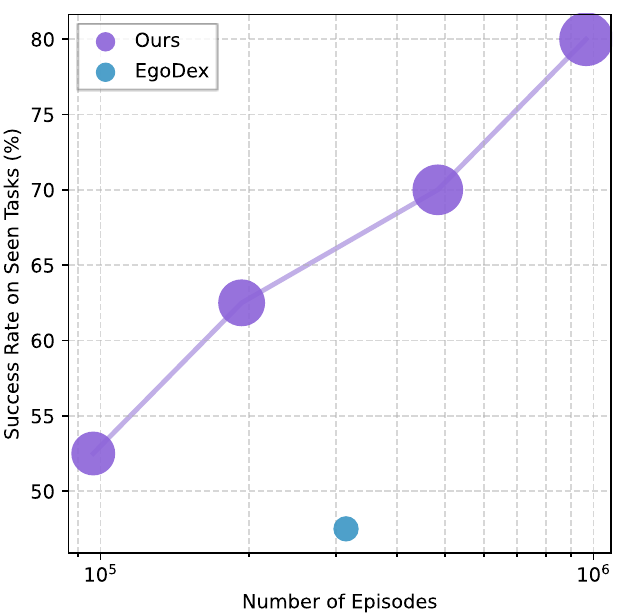}
    \captionsetup{margin={0cm,-0.5cm}}
        \caption{}
        \label{fig:data_scale_robot_seen}
    \end{subfigure}
    \hfill
    \begin{subfigure}[t]{0.3\linewidth}
        \centering
        \includegraphics[height=5cm]{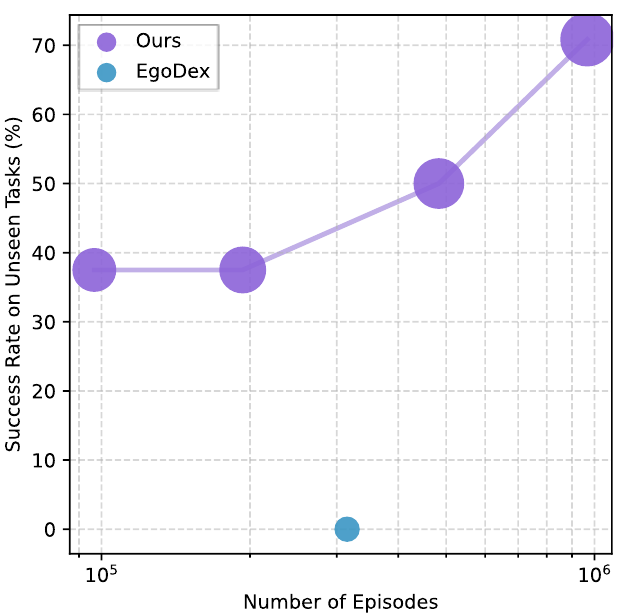}
    \captionsetup{margin={0cm,-0.5cm}}
        \caption{}

        \label{fig:data_scale_robot_unseen}
    \end{subfigure}
    \hfill
    \begin{subfigure}[t]{0.34\linewidth}
        \centering
        \includegraphics[height=5cm]{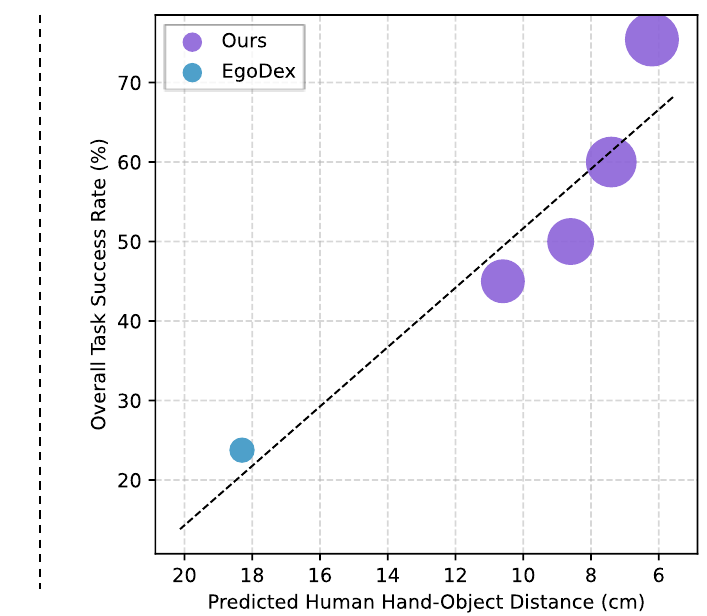}
    \captionsetup{margin={0cm,-1.1cm}} 
    \caption{}
        \label{fig:data_scale_robot_correlation}
    \end{subfigure}
    \vspace{-4pt}
    \caption{Data scaling behavior on real-robot pick-and-place tasks. The circle size indicates the \emph{visual diversity} of the pretraining data. (a) Task success rate on \emph{seen} objects and backgrounds. (b) Task success rate on \emph{unseen} objects and backgrounds. (c) Correlation between robot task performance and pretraining hand action prediction accuracy.}
    \label{fig:data_scale_robot}
\end{figure*}

Results of different methods are presented in Tab.~\ref{tab:robot},~\ref{tab:robot_unseen}, and Fig.~\ref{fig:data_scale_robot}. Compared to the model without human VLA data pretraining, our approach achieves superior execution success and stronger generalization on unseen tasks. Compared to OXE pretraining, our method achieves substantially stronger few-shot and unseen task performance. The OXE dataset contains data from gripper-based robots and offers far less diversity in objects, tasks, and environments compared to our human hand VLA dataset, as discussed in Sec.~\ref{sec:dataset}. Moreover, the significant differences among its various sub-datasets also reduce the model's generalization ability due to shortcut learning~\citep{xing2025shortcut}. EgoDex pretraining leverages carefully collected large-scale human-hand data, yet its limited environmental diversity leads to lower success rates and reduced generalization performance. As shown in Fig.~\ref{fig:data_scale_robot}, EgoDex performs worse than our model pretrained on only 10\% of the data, even though it contains more episodes and a significantly larger number of frames (\ie, 130M \emph{v.s.} 2.6M). Moreover, it completely fails on unseen scenes, highlighting the importance of data diversity for generalization. 

\paragraph{Influence of Action Representation} We further compare with a baseline method, \emph{Latent action pretrain}, to validate the effectiveness of pretraining with explicit 3D action predictions. For this baseline, we use the same episode splits and language instructions, but replace the original 3D actions to be predicted with latent actions from LAPA~\citep{yelatent2025} trained on our data. 

The comparison results are also presented in Tab.~\ref{tab:robot} and~\ref{tab:robot_unseen}. As shown, while latent action pretraining performs moderately on seen tasks, it fails completely in unseen environments. This is likely due to latent actions struggling to disentangle task-relevant motions from task-irrelevant background, causing them to fail in novel settings\footnote{Some recent approaches~\citep{bu2025univla} focus on better disentangling task-irrelevant information in latent actions. We leave a comparison with these methods for future work.}. By contrast, our approach achieves significantly better performance, benefiting from more explicit action supervision, which leads to a smaller pretraining–finetuning gap. This demonstrates the advantage of closely aligning human video data with robotic VLA data during pretraining.

\paragraph{Data Scaling Behavior} Following the human-hand prediction experiments, we examine how the scale of pretraining data affects robotic task performance. For this experiment, we compare with models pretrained on human-hand data using 50\%, 20\%, and 10\% of the dataset (we do not include the 1\% case, as we consider this amount too small for effective pretraining before fine-tuning). We focus on general pick-and-place tasks involving both seen and unseen objects (categories) and backgrounds.
Figure~\ref{fig:data_scale_robot} illustrates the relationship between the scale of pretraining data and robotic task success rates, showing consistent improvements on both seen and unseen tasks as the data scale increases.

\paragraph{Robot Performance \emph{v.s.} Pretraining Hand-Prediction Accuracy} Finally, we investigate the relationship between the fine-tuned robotic task success rates and the pretraining accuracy on human-hand prediction. The former is evaluated by the average success rate across seen and unseen pick-and-place tasks, while the latter is measured by the hand–object distance reported in Tab.~\ref{tab:pretrain}. As shown in Fig.~\ref{fig:data_scale_robot_correlation}, the two exhibit a clear positive correlation: models that achieve higher performance on the human-hand prediction benchmark also yield higher task success rates after fine-tuning on robot data. This suggests that our hand action prediction benchmark can serve as an effective proxy for downstream robotic performance, enabling rapid prototyping of pretrained VLA models.

%% file: sections/5_discussion.tex
\section{Discussion and Future Work}
Our method serves as an initial exploration toward constructing large-scale VLA pretraining data from real-life human activity videos. While our current data are mainly sourced from existing egocentric human video datasets, the automatic data pipeline is readily extensible. In future work, we plan to incorporate more diverse video sources (\eg, Howto100M~\citep{miech2019howto100m}) to enable larger-scale and more comprehensive VLA pretraining. Due to the limitations of current 3D reconstruction algorithms and the inherent capabilities of the VLM, our constructed pretraining data still exhibits some inaccuracies. We aim to further improve its quality using more advanced reconstruction techniques, and introduces additional filtering mechanisms to remove noisy samples. In addition, our present data construction and model training mainly target short-horizon, atomic manipulation skills. Extending this framework to organize data into higher-level task structures for learning long-horizon planning and reasoning abilities represents an important future direction.

Our current robotic experiments primarily focus on single-handed manipulation tasks. Nevertheless, our framework naturally supports bimanual operations. We conduct a simple “hand-over” experiment to demonstrate its feasibility on two-handed tasks, as shown in the bottom row of Fig.~\ref{fig:long_exc}. We plan to investigate more bimanual scenarios in future work. Furthermore, integrating our pretraining framework with multi-view visual inputs and tactile feedback to handle more complex manipulation tasks is another promising direction for exploration.

%% file: sections/suppl.tex
\section{More Implementation Details} \label{sec:more_implement}

\subsection{Hand V-L-A Data Construction} \label{sec:more_implement_data}

\subsubsection{3D Motion Labeling} 

\paragraph{Camera Intrinsics Estimation} For videos captured with moving cameras, we apply DroidCalib to estimate the camera intrinsics under the unified camera model~\citep{mei2007single}, which extends the conventional pinhole model by introducing an additional distortion parameter to better handle ultra-wide-angle and fisheye cameras. During estimation, we assume the principal point lies at the image center and the focal lengths are identical in both image axes. For static cameras, we first employ DeepCalib to estimate intrinsics under the same unified camera model assumption. If the distortion coefficients are small, we apply MoGe-2 to estimate the final focal length under the pinhole camera model assumption. For videos with non-negligible distortion, an additional undistortion step is applied to make the images conform to the pinhole model before subsequent processing.

\paragraph{Video Hand Reconstruction} We employ HaWoR for camera-space 3D hand reconstruction which is capable of jointly reconstructing hands within each video chunk. The camera focal information estimated in the previous stage is provided as input to HaWoR to facilitate accurate hand reconstruction. The original HaWoR includes a hand motion infiller module that interpolates missing frames in the 3D reconstruction. We discard this module in our pipeline, as its interpolation results are less reliable on in-the-wild videos.

\paragraph{Camera Pose Estimation} We use MegaSAM for metric-scale camera pose estimation. The original MegaSAM incorporates depth priors predicted by DepthAnything~\citep{yang2024depth} and UniDepth~\citep{piccinelli2024unidepth} to address challenges in videos with limited camera baselines and complex scene dynamics.
In our initial exploration, we found that replacing these depth modules with direct outputs from MoGe-2 yields more accurate and stable results, while significantly improving inference efficiency. Therefore, we adopt this modified version of MegaSAM in our framework. Additionally, when estimating the camera, we initialize MegaSAM’s camera intrinsics using the focal length information obtained from the previous stage.

\subsubsection{Atomic Action Segmentation} We segment long videos into atomic-level video clips using speed minima of the 3D hand wrists in the world space. Before computing the minima, the 3D wrist trajectories are smoothed with a Gaussian filter in world space to mitigate spurious extrema caused by reconstruction noise. Additionally, a detected minimum is required to be the smallest value within a 0.5-seconds window centered on it to further mitigate noise effects.

\subsubsection{Instruction Labeling} \label{sec:more_implement_data_gpt}
\begin{figure*}[p]
    \centering
    \includegraphics[width=1.0\linewidth]{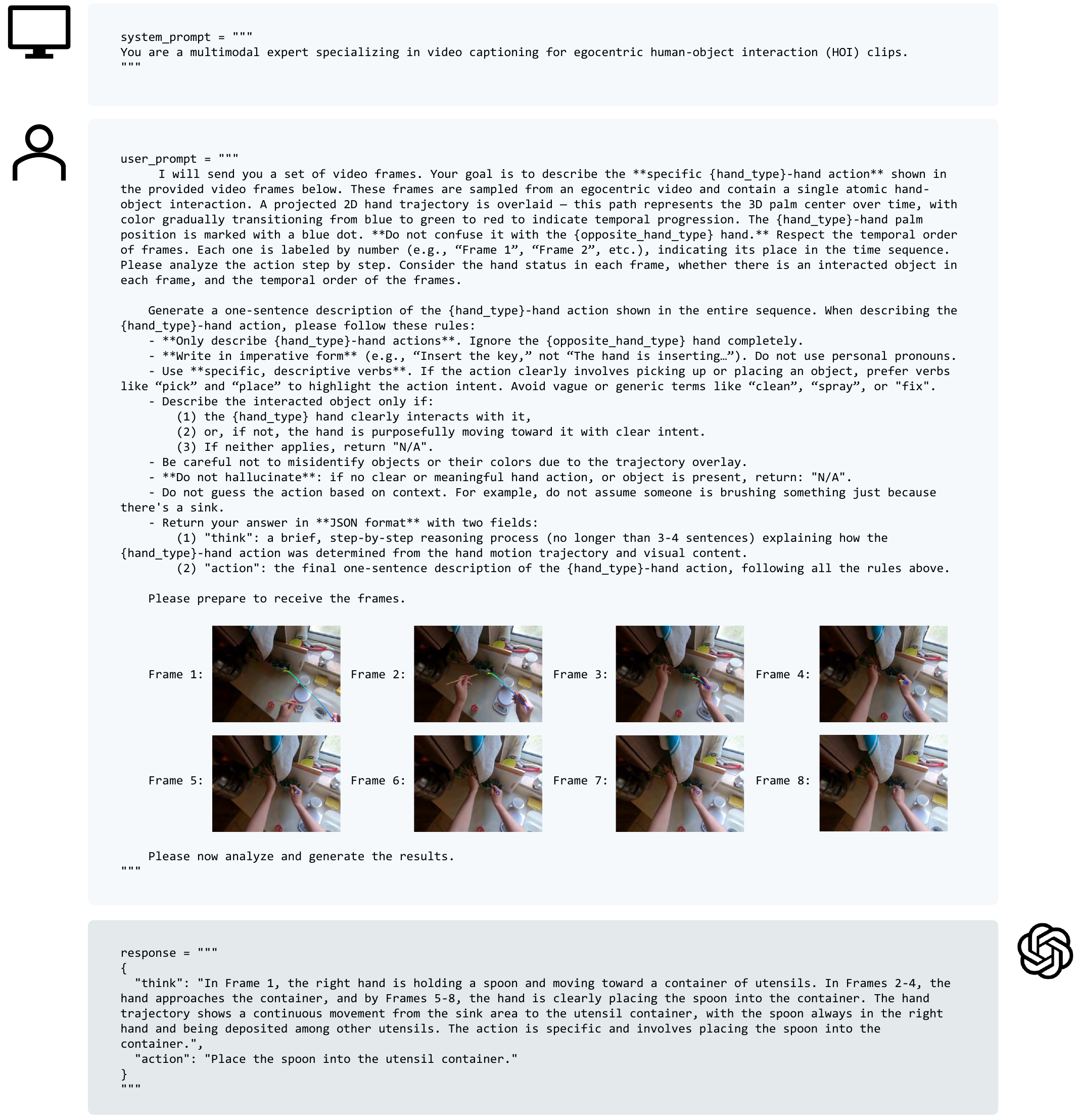}
    \caption{An example of our instruction labeling process.}
    \label{fig:gpt_prompt}
\end{figure*}

We prompt GPT-4.1 to generate action captions based on sampled frames from atomic-level video clips. A detailed example is illustrated in Fig.~\ref{fig:gpt_prompt}. After obtaining a caption for a video clip, we further ask GPT-4.1 to rephrase it into five diverse versions while preserving the original meaning, in order to increase the diversity of language descriptions.

\subsection{Model Architecture} \label{sec:more_implement_model}
\subsubsection{VLM Backbone} 
Our VLM is based on the 3B version of PaliGemma-2 with ``mix checkpoint'' that has been further fine-tuned on multiple downstream tasks. The input image resolution is $224^2$, and all images of varying sizes are directly resized to this resolution without any additional center cropping. To improve the model’s understanding of the original images’ aspect ratio and camera intrinsics, we incorporate the camera FoV as an additional token. The FoV token is projected via an MLP to align with the embedding space of the LLM input tokens, allowing the model to interpret the geometric characteristics of visual inputs more effectively. This is important because the SigLIP encoder processes images with a fixed 1:1 aspect ratio, which may lead to ambiguity if FoV cues are absent. Moreover, our trajectory-aware augmentation (see Sec.~\ref{sec:pretrain}) modifies the FoV, necessitating the model to consider these variations for correct interpretation of the augmented images.

\subsubsection{Diffusion Action Expert} 
The action expert is implemented as a Diffusion Transformer with approximately 136M parameters. Within each transformer block, we replace bidirectional self-attention with causal self-attention for action tokens (see Sec.~\ref{sec:causal} for further discussion), and QKNorm~\citep{henry2020query} is applied in the attention layers while LayerNorm is replaced with RMSNorm~\citep{zhang2019root} to improve training stability. The cognition feature $\bm f^c$, the hand state $\bm s_t$, and the noisy action chunk are first projected via an MLP and subsequently processed through a causal self-attention layer. Additionally, $\bm f^c$ is injected via AdaLN to further enhance vision–language conditioning. 
\subsubsection{State and Action Normalization}
For both state and action inputs to the action expert, we apply mean–variance normalization to each dimension, standardizing them to zero mean and unit variance.
During pretraining, dataset-specific statistics are first computed, after which unified normalization parameters are derived by weighting each dataset according to its frame sampling probability.
These parameters are then kept fixed throughout pretraining and zero-shot hand action prediction evaluation. During fine-tuning, we recompute the mean and variance from the robot data and apply normalization accordingly.

\subsection{Training Details} \label{sec:more_implement_train}
During training, we employ PyTorch's Fully Sharded Data Parallel (FSDP) framework. The length of the action chunk is set to 16, and the diffusion process uses 100 noise steps. For each VLM forward pass, eight noisy samples are randomly drawn to train the action expert efficiently, following~\citep{li2024cogact}. AdamW~\citep{loshchilov2017decoupled} is used as the optimizer with a weight decay of 1e-1 and a gradient clipping value of 1.0, applied consistently in both pretraining and fine-tuning stages. The $\beta$ parameters of AdamW are set to $(\beta_1, \beta_2) = (0.9, 0.99)$ during pretraining and $(\beta_1, \beta_2) = (0.9, 0.95)$ during fine-tuning to enhance training stability.

At the pretraining stage, we employ trajectory-aware augmentation as discussed in Sec.~\ref{sec:pretrain}. For episodes annotated with multiple language instructions by GPT (see Appendix~\ref{sec:more_implement_data_gpt}), a single instruction is randomly selected per trajectory. The state input $\bm s_t$ to the action expert is dropped with a probability of 0.1, encouraging the model to rely solely on vision–language input and preventing overfitting to the state. Similarly, the cognition token is dropped with a probability of 0.1 in the action expert to leverage classifier-free guidance (CFG)~\citep{ho2022classifier}. 

During finetuning, we align the coordinate systems of the real-robot and pretrained human data, and map the robot hand’s action dimensions to match those of the human hand, following the discussion in Sec~\ref{sec:finetune}. By default, the pretrained model is finetuned for 20K steps. The model without VLA pretraining are finetuned for 60K steps, as they do not converge within 20K steps and produce highly jittery actions. For other configurations, including ablations and prior methods, we select the best-performing checkpoint every 10K steps. All prior methods used for comparison are implemented using their official repositories.

\subsection{Inference Details}
During inference, we use DDIM~\citep{song2020denoising} with 10 sampling steps and a CFG scale of 5.0. For real-robot execution, we adopt the action chunking strategy~\citep{zhao2023learning,li2024cogact}, executing 6 out of 16 actions at a time. Predicted end-effector actions in the camera coordinate frame are first converted to absolute 6D poses in the robot coordinate frame, then transformed into joint angles using an inverse kinematics (IK) solver. The resulting hand joint angles are directly mapped to the corresponding robot dexterous hand joints for execution. Experiments for \emph{Human Hand Action Prediction} (Sec.~\ref{humanhandprediction}) are conducted on a single NVIDIA A6000 GPU, while \emph{Real-World Robot Dexterous Manipulation} experiments (Sec.~\ref{sec:exp_robot}) use a single NVIDIA 4090 GPU.

\subsection{Robot Teleoperation System} \label{sec:more_implement_robot}

\subsubsection{Leader–Follower Teleoperation Arm} We use a leader–follower arm system for teleoperation data collection. The leader arms share the same joint topology as the Realman robot arms. Operators manipulate the leader arms to perform desired motions, and the joint angles measured from the leader arms are directly sent to the follower arms (\ie, robot arms) to replicate these motions, enabling precise control of the end-effector 6D pose.

\subsubsection{Hand Pose Retargeting}
We use MANUS gloves to directly control the dexterous robot hand and employ a hand pose retargeting method to map the glove measurements to the joint angles of the robot hand for precise motion control. At each timestep, the retargeting algorithm optimizes the mapping to transform human hand motions into the corresponding robot hand movements. We implement two optimizers in our pipeline, as described below:

\paragraph{DexPilot Optimizer}
The first follows the approach of DexPilot~\citep{handa2020dexpilot} and the implementation of AnyTeleop~\citep{qin2023anyteleop}. We define a set of vectors $\mathcal{V}$ consisting of the five wrist-to-fingertip vectors and ten inter-finger vectors. The objective is to minimize the squared difference between the glove keypoint vectors $v_i^h$ and the corresponding robot vectors $v_i^r(q_t)$ obtained through forward kinematics:
\begin{equation}
    \mathcal{L}_{\text{vec}}(q_t) = \sum_{i=0}^{N} s(d_i) \left\| \alpha v_i^h - v_i^r(q_t) \right\|^2 + \beta \left\| q_t - q_{t-1} \right\|^2,
    \label{eq:vec_eq}
\end{equation}
subject to $q_l \leq q_t \leq q_u$, where $q_l$ and $q_u$ denote the joint limits of the robot hand, $\alpha$ is a scaling factor to account for different hand sizes, and $\beta$ is a weight for temporal smoothness. The switching weight function $s(d_i)$ increases as the distance $d_i$ between the fingertip and wrist decreases, encouraging fingertip contact.

\paragraph{Angle Matching}
The second optimizer focuses on two key vectors: the thumb–wrist vector and the vector connecting the index fingertip to its root. Their optimization follows the same formulation as Eq.~\eqref{eq:vec_eq}. Notably, only the lateral-swing degrees of freedom (\ie, abduction and adduction) of the thumb and index finger are updated based on this optimization, while their flexion degrees of freedom are excluded. All remaining joints are directly controlled using angles derived from the glove keypoints.

Given glove keypoints $\mathbf{k}_i \in \mathbb{R}^3$, joint angles are computed from triplets $(A_j,B_j,C_j)$ as
\[
\theta_j^h = \arccos\!\left(\frac{(\mathbf{k}_{A_j}-\mathbf{k}_{B_j})^\top(\mathbf{k}_{C_j}-\mathbf{k}_{B_j})}
{\|\mathbf{k}_{A_j}-\mathbf{k}_{B_j}\|\,\|\mathbf{k}_{C_j}-\mathbf{k}_{B_j}\|}\right),
\]
with the sign determined by the cross product relative to a reference axis, where $(A_j,B_j,C_j)$ denote the indices of three glove keypoints forming two connected bone segments.  
Each glove angle $\theta_j^h$ is then linearly mapped from its empirical range 
$[\theta^{h,\min}_j,\theta^{h,\max}_j]$ to the corresponding robot joint limits 
$[\theta^{r,\min}_j,\theta^{r,\max}_j]$:
\[
\theta^{r}_j = 
\frac{\theta_j^h - \theta^{h,\min}_j}{\theta^{h,\max}_j - \theta^{h,\min}_j}
\big(\theta^{r,\max}_j - \theta^{r,\min}_j\big) + \theta^{r,\min}_j,
\]
and clipped to the valid range $[\theta^{r,\min}_j,\theta^{r,\max}_j]$.  
The resulting $\theta^r$ values are applied directly as angle commands to the robot hand.

\section{More Evaluation Details}  \label{sec:more_eval}

\subsection{Hand Action Prediction Benchmark} We use the benchmark described in Sec.~\ref{sec:hand_benchmark} to quantitatively evaluate the performance of the pretrained VLA models on the grasping task. The RGB-D images captured with the Azure Kinect are first undistorted to remove camera distortion. Then, for each image, we manually annotate the object positions and their corresponding captions. The annotated object positions are subsequently used as prompts for SAM-2~\citep{ravi2024sam} to obtain the corresponding object masks. These masks, together with the depth images and the camera intrinsics, are used to reconstruct the 3D point cloud of each object in the camera coordinate frame.

For each image-caption pair corresponding to a target object in the scene, we render a synthetic hand with an attached arm using SMPL-X~\citep{pavlakos2019expressive}, a parametric full-body model that incorporates the MANO hand. We assign the hand a natural resting pose and place it approximately 20 cm away from the target object. We ensure that the synthetic hand is always positioned closer to the camera than the target object, preventing incorrect occlusion relationships in the rendered images. The images with the rendered hand, along with the corresponding language instructions, are then used to evaluate VLA hand action prediction. We prompt the model to generate a single action chunk, which corresponds to approximately 0.5 s of motion for our model and 1 s for Being-H0, and compute the minimal distance between the fingertip positions and the object’s point cloud. For each instruction, we independently generate four trajectories and report the mean of their minimal hand–object distances. The final score is obtained by averaging (or taking the median of) these distances across all image–instruction pairs.

\subsection{User Study} 
\begin{figure*}[p]
    \centering
    \includegraphics[width=1.0\linewidth]{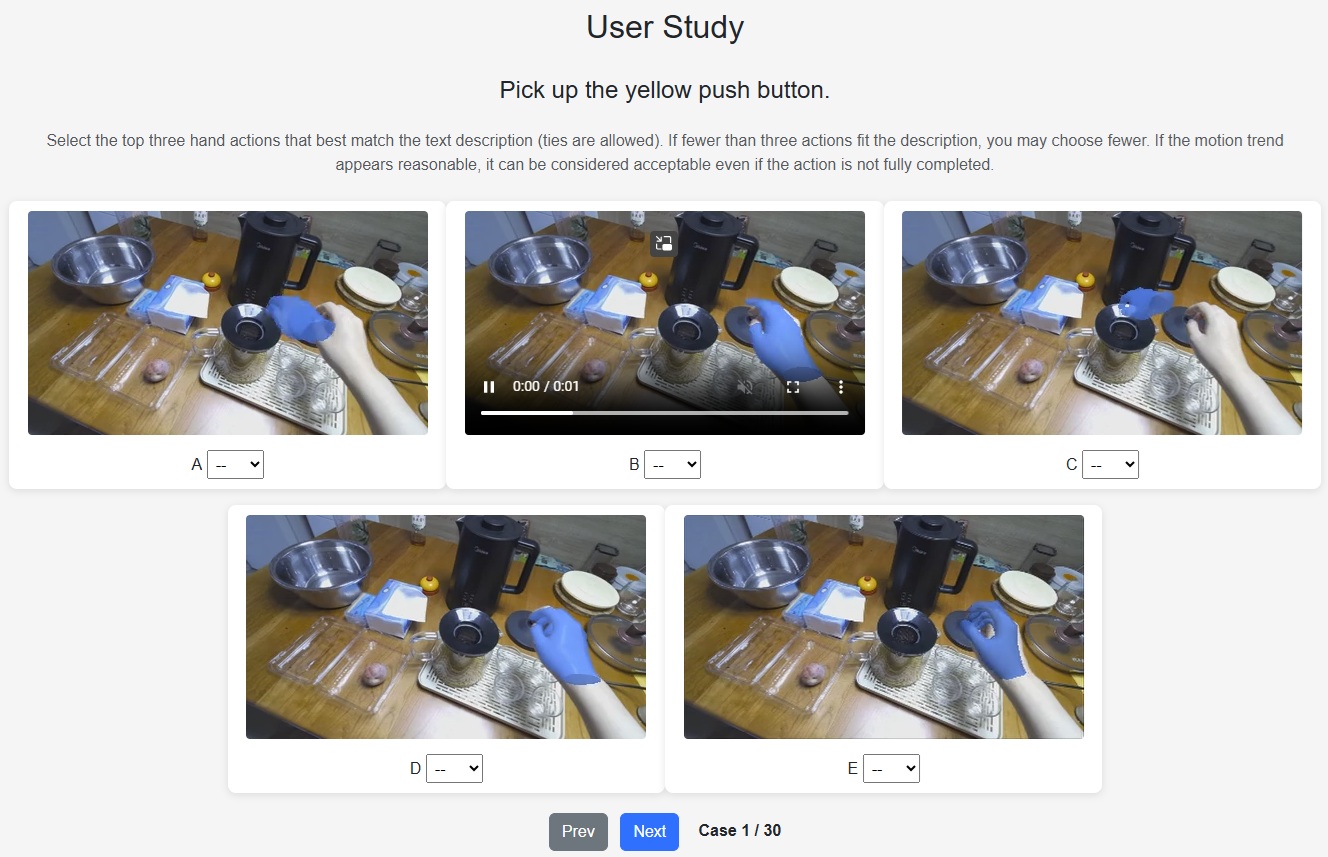}
    \caption{User study interface for hand action prediction.}
    \label{fig:user_study_interface}
\end{figure*}

For more general hand actions, we assess the plausibility of the action predictions through a user study. We developed a website to allow participants to evaluate the quality of actions generated by different methods. The user interface is shown in Fig.~\ref{fig:user_study_interface}. Each participant was assigned 30 trials, randomly selected from a total of 117 scenes. For each case, the results from different methods were anonymized and presented in a randomized order to avoid any bias. The hand motions generated by each method were rendered as videos, and participants were asked to judge whether the generated video matched the given language instruction and to rank the different methods accordingly.

\section{More Results}  \label{sec:more_results}

\subsection{Human Hand V-L-A Data}  \label{sec:more_results_data}
\begin{figure*}[p]
    \centering
    \includegraphics[width=1.0\linewidth]{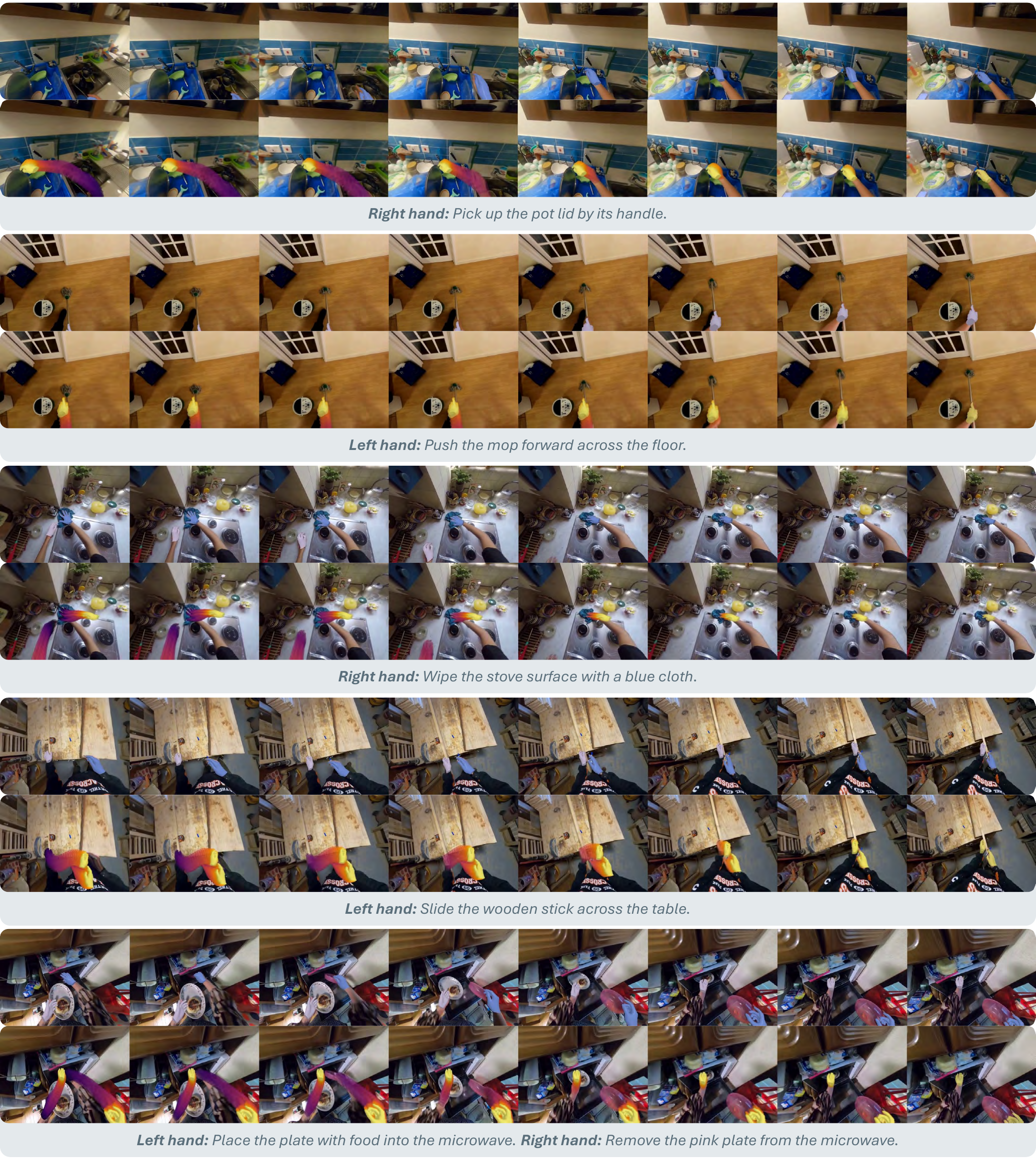}
    \caption{Examples of our hand V-L-A data used for pretraining. The first row in each case visualizes the reconstructed 3D hands for individual frames, while the second row shows the hand action trajectory from the current frame to the end of the episode. The color gradient from purple to yellow indicates the temporal progression from the beginning to the end of the episode.}
    \label{fig:pretrain_data1}
\end{figure*}

\begin{figure*}[p]
    \centering
    \includegraphics[width=1.0\linewidth]{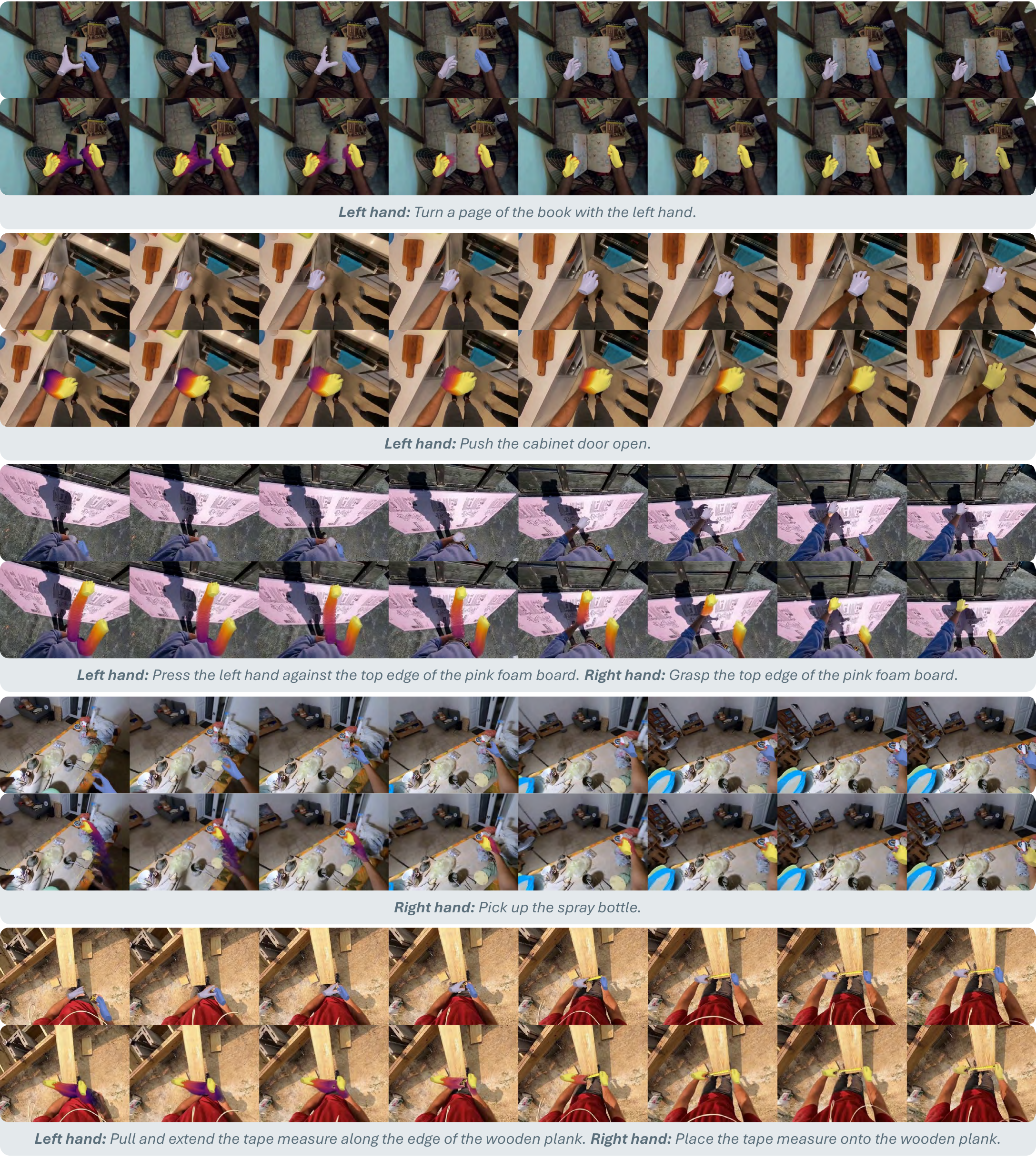}
    \caption{Examples of our hand V-L-A data used for pretraining. The first row in each case visualizes the reconstructed 3D hands for individual frames, while the second row shows the hand action trajectory from the current frame to the end of the episode. The color gradient from purple to yellow indicates the temporal progression from the beginning to the end of the episode.}
    \label{fig:pretrain_data2}
\end{figure*}

We showcase additional hand VLA data constructed using our method in Figs.~\ref{fig:pretrain_data1} and~\ref{fig:pretrain_data2}. Our dataset covers a wide range of environments and hand actions. Moreover, due to the presence of moving cameras in our scenes, the observations across different frames show noticeable variation, further increasing the diversity of the data.

\subsection{Hand Action Prediction Results} \label{sec:more_results_hand}
\begin{figure*}[p]
    \centering
    \includegraphics[width=1.0\linewidth]{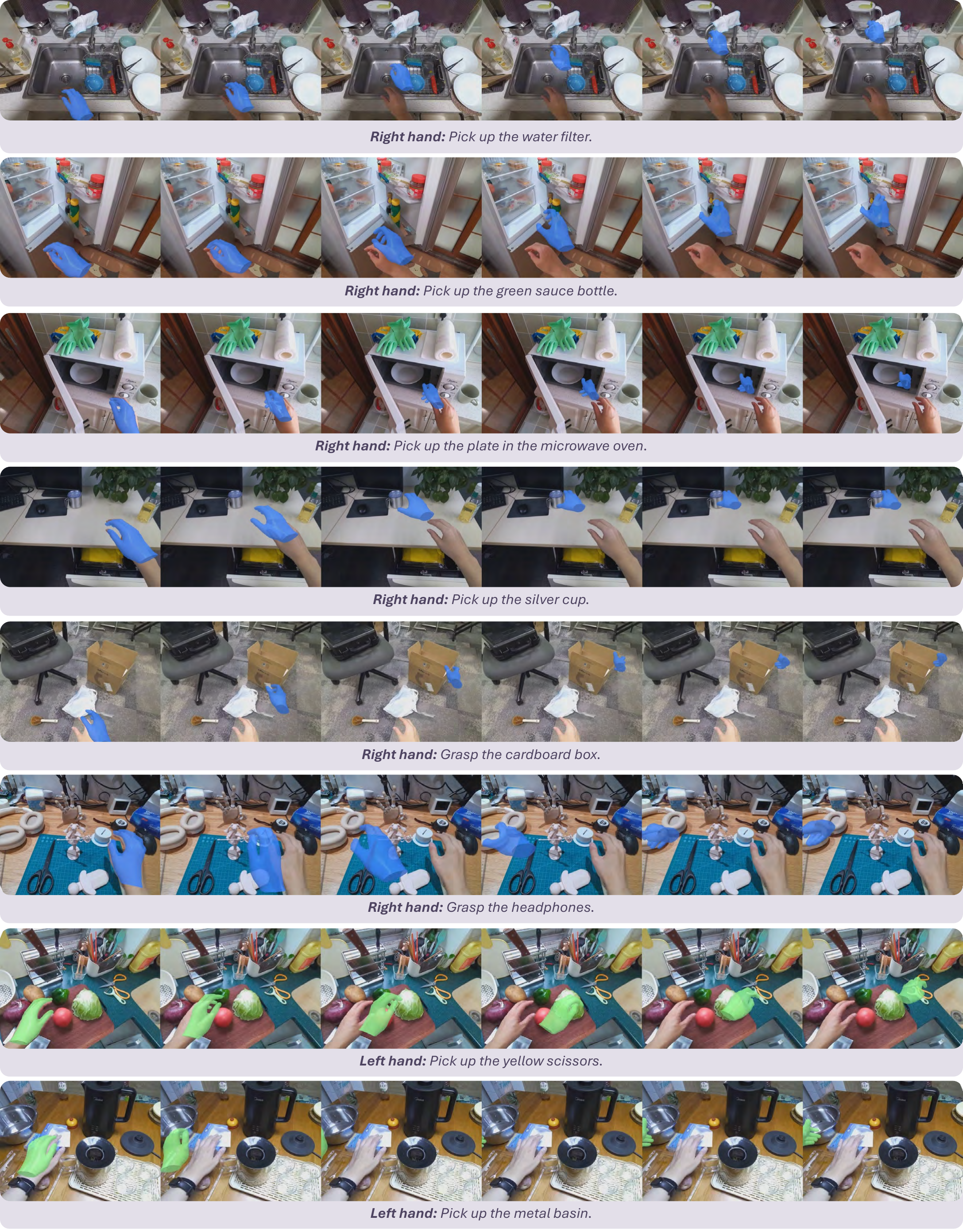}
    \caption{Hand action prediction (grapsing) in unseen real-world environments by our method, with time increasing from left to right.}
    \label{fig:hand_action_pick1}
\end{figure*}

\begin{figure*}[p]
    \centering
    \includegraphics[width=1.0\linewidth]{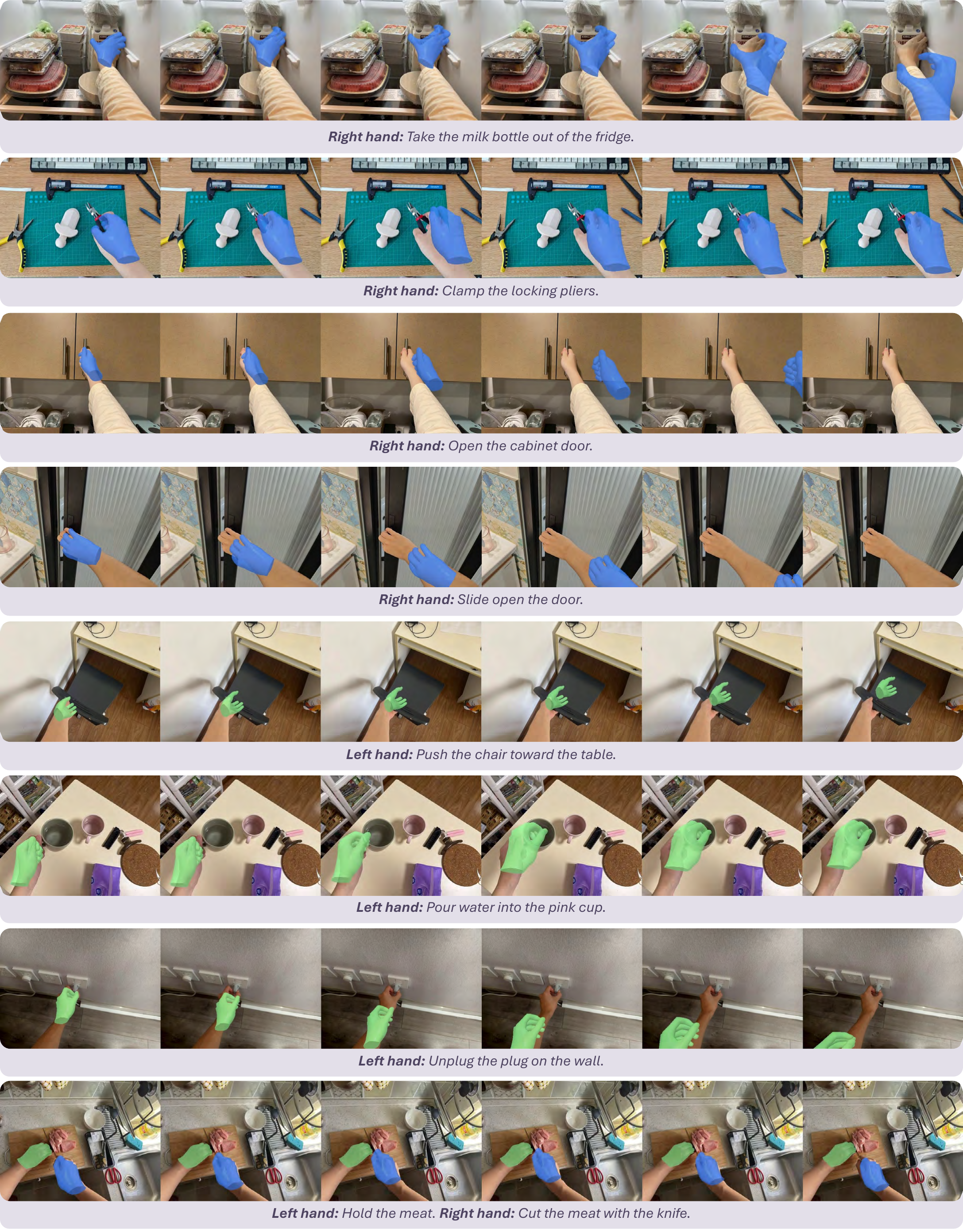}
    \caption{Hand action prediction (general action) in unseen real-world environments by our method, with time increasing from left to right.}
    \label{fig:hand_action_pick2}
\end{figure*}

We provide more hand action prediction results on \emph{unseen real-life environments} in Fig.~\ref{fig:hand_action_pick1} and~\ref{fig:hand_action_pick2}. Our pretrained VLA model demonstrates strong generalization across these scenes and is capable of predicting diverse human hand motions.

\subsection{Real-Robot Execution Results} \label{sec:more_results_robot} 
\begin{figure*}[p]
    \centering
    \includegraphics[width=1.0\linewidth]{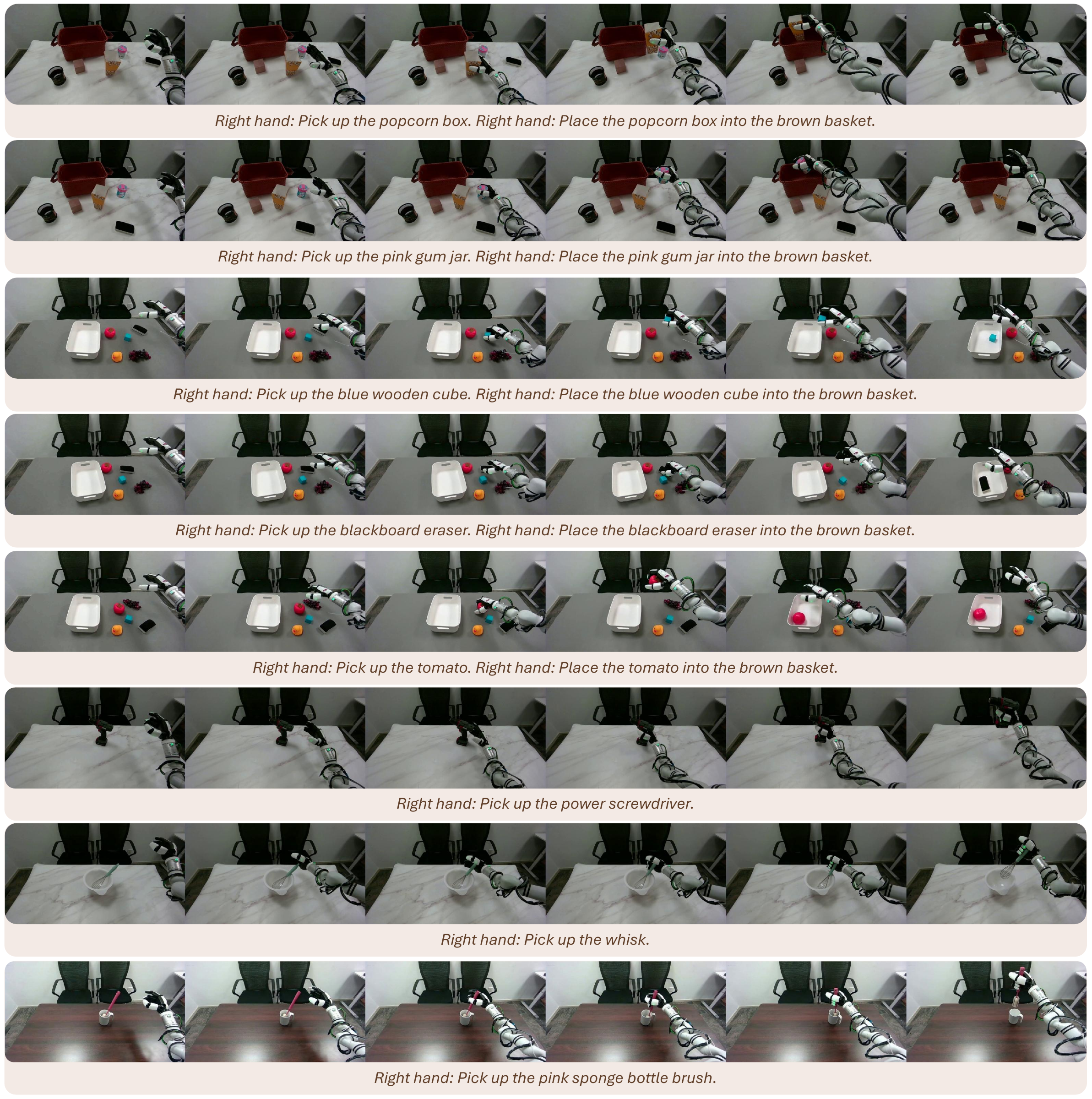}
    \caption{In-domain execution trajectories of the \textit{general pick-and-place} task and the \textit{functional grasping} task.
Rows 1--5 show examples of execution trajectories for the \textit{general pick-and-place} task, while rows 6--8 present examples for the \textit{functional grasping} task. Images are captured with the robot head camera, with time increasing from left to right.}
    \label{fig:seen_task_exc}
\end{figure*}

\begin{figure*}[p]
    \centering
    \includegraphics[width=1.0\linewidth]{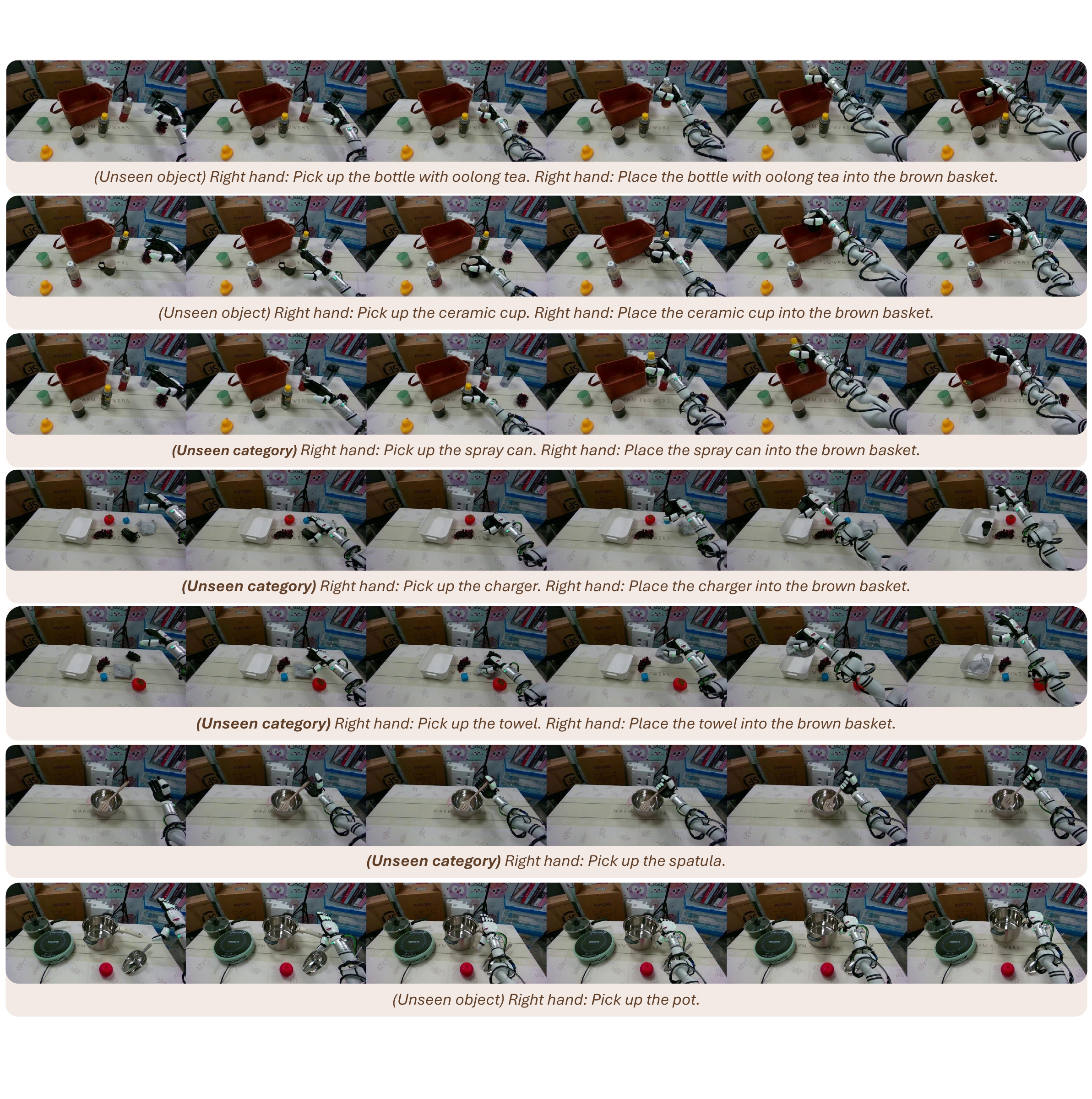}
    \vspace{-36pt}
    \caption{Execution trajectories of the \textit{general pick-and-place} task and the \textit{functional grasping} task with \textit{unseen background and objects}.
Rows 1--5 show examples of execution trajectories for the \textit{general pick-and-place} task, while rows 6--7 present examples for the \textit{functional grasping} task. \textit{Unseen object} means
the objects are new but other objects of the same categories were seen in fine-tuning; and \textit{unseen categories} means the objects belong to categories not encountered before. Images are captured with the robot head camera, with time increasing from left to right.}
    \label{fig:unseen_exc}
\end{figure*}

\begin{figure*}[p]
    \centering
    \includegraphics[width=1.0\linewidth]{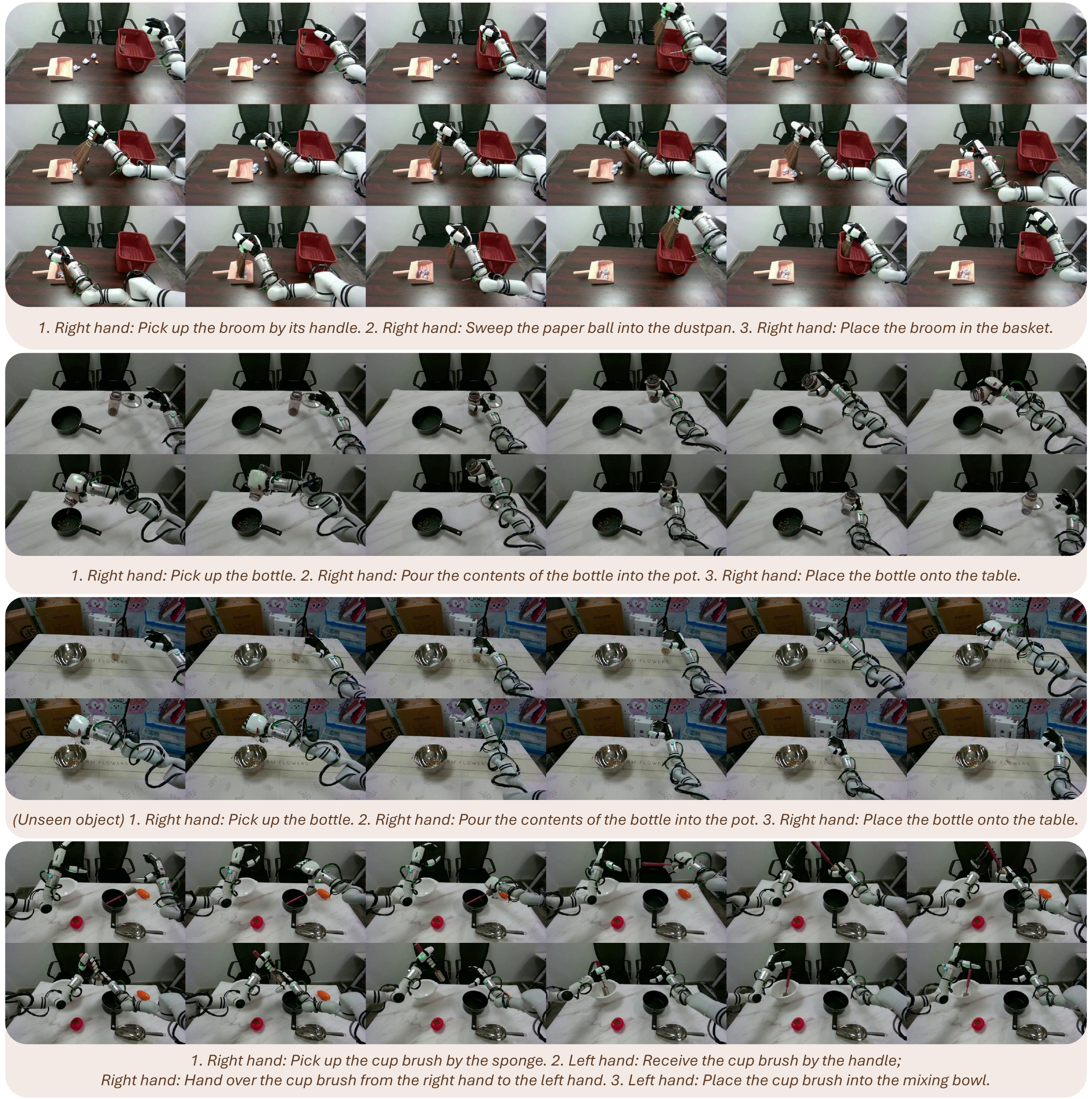}
    \caption{Execution trajectories of the sequential tasks. 
Rows 1--3 correspond to the \textit{sweeping} task, rows 4--5 correspond to the \textit{seen pouring} task, rows 6--7 correspond to the \textit{pouring} task with \textit{unseen backgrounds and objects}, and rows 8--9 correspond to the \textit{bimanual dexterous handover} task. 
The numbers indicate the execution order of these tasks. Images are captured with the robot head camera, with time increasing from left to right and from top to bottom.}
    \label{fig:long_exc}
\end{figure*}

Figures~\ref{fig:seen_task_exc},~\ref{fig:unseen_exc}, and~\ref{fig:long_exc} present additional visual results of real robot task executions in our experiments. In the last row of Fig.~\ref{fig:long_exc}, we additionally showcase a bimanual ``hand over'' task, demonstrating the framework’s ability to transfer naturally to bimanual tasks.

